\documentclass[journal]{IEEEtran}
\usepackage{amsmath,amssymb}
\usepackage{overpic}
\usepackage{hyperref}
\hypersetup{
	colorlinks=true,
	linkcolor=black,
	filecolor=black,
	urlcolor=black,
	citecolor=black,
}
\usepackage{booktabs}
\usepackage{multirow}
\usepackage{color}
\usepackage{cite}
\newcommand{\PLH}{{\mkern0mu\times\mkern0mu}}

\begin{document}
\title{Monocular 3D Object Detection with Sequential Feature Association and Depth Hint Augmentation}
\author{Tianze Gao, ~Huihui Pan ~and ~Huijun Gao, ~\IEEEmembership{Fellow,~IEEE}
	\thanks{Manuscript received December 1, 2020; revised June 18, 2021 and November 16, 2021; accepted January 2, 2022.  This work was supported in part by the National Natural
Science Foundation of China under Grant U1964201, Grant 62173108; in part by the Major Scientific and Technological Special
Project of Heilongjiang Province under Grant 2021ZX05A01; and in part by XPLORER PRIZE. (Corresponding
authors: Huihui Pan; Huijun Gao.)}
	\thanks{T. Gao is with the Research Institute of Intelligent Control and Systems, Harbin Institute of Technology, Harbin, 150001, China, and also with the Ningbo Institute of Intelligent Equipment Technology Co., Ltd., Ningbo 315200, China (e-mail: gao2990026796@gmail.com).}
	\thanks{H. Pan is with the Research Institute of Intelligent Control and Systems, Harbin Institute of Technology, Harbin, 150001, China, and also with the Harbin Institute of Technology Robot Innovation Center Co., Ltd., Harbin, 150001, China. (e-mail: huihuipan@hit.edu.cn).}
	\thanks{H. Gao is with the Research Institute of Intelligent Control and Systems, Harbin Institute of Technology, Harbin, 150001, China, and also with the Department of Mathematics and Theories, Peng Cheng Laboratory, Shenzhen 518055, China. (e-mail: hjgao@hit.edu.cn).}
	\thanks{Color versions of one or more figures in this article are available at https://doi.org/10.1109/TIV.2022.3143954.}
	\thanks{Digital Object Identifier 10.1109/TIV.2022.3143954}
}
\maketitle

\begin{abstract}
Monocular 3D object detection, with the aim of predicting the geometric properties of on-road objects, is a promising research topic for the intelligent perception systems of autonomous driving. Most state-of-the-art methods follow a keypoint-based paradigm, where the keypoints of objects are predicted and employed as the basis for regressing the other geometric properties. In this work, a unified network named as FADNet is presented to address the task of monocular 3D object detection. In contrast to previous keypoint-based methods, we propose to divide the output modalities into different groups according to the estimation difficulty of object properties. Different groups are treated differently and sequentially associated by a convolutional Gated Recurrent Unit. Another contribution of this work is the strategy of depth hint augmentation. To provide characterized depth patterns as hints for depth estimation, a dedicated depth hint module is designed to generate row-wise features named as depth hints, which are explicitly supervised in a bin-wise manner. The contributions of this work are validated by conducting experiments and ablation study on the KITTI benchmark. Without utilizing depth priors, post optimization, or other refinement modules, our network performs competitively against state-of-the-art methods while maintaining a decent running speed.
\end{abstract}


\section{Introduction}
\label{sec:introduction}
\IEEEPARstart{M}{onocular} 3D object detection in autonomous driving scenarios, with the goal of estimating the 2D and 3D geometric properties of on-road objects in a monocular image, is a challenging yet promising research topic. The state-of-the-art 3D detection methods~\cite{shi2019pointrcnn,yang2019std,yan2018second,rozsa2019object,yang20203dssd} generally resort to the power of Light Detection and Ranging (LiDAR) sensors. Despite that these LiDAR-based methods can achieve high performance, they often require excessive hardware expense. As compared, the monocular 3D object detection problem, which is the focus of this work, simply requires a single RGB image as input.

Detecting the 3D geometric information (3D center location, orientation, dimension, etc.) of on-road objects by using monocular vision is inherently an ill-posed problem, since the depth values of 3D points become computationally unrecoverable once they are projected onto 2D image planes. However, by exploiting the appearance, surrounding context, and the empirical dimensions of objects, artificial neural networks (ANN) are capable of producing reasonable predictions about objects' 3D properties.

Currently, the majority of monocular 3D object detection methods is dependent on region proposals to seek for potential object candidates~\cite{chen2016monocular,xiang2017subcategory,mousavian20173d,chabot2017deep,bao2019monofenet,he2019mono3d++,hu2019joint,manhardt2019roi,ku2019monocular,brazil2019m3d}. The region proposals are either generated under the guidance of multiple priors~\cite{chen2016monocular}, by using a region proposal network~\cite{chabot2017deep,xiang2017subcategory,mousavian20173d,bao2019monofenet,manhardt2019roi,ku2019monocular}, via trained 2D detectors~\cite{chabot2017deep,he2019mono3d++,hu2019joint} or being embedded into a single-stage pipeline~\cite{brazil2019m3d}. As an alternative, the task of 3D object detection can also be phrased as the combination of keypoint detection and class-agnostic regression. Previous works~\cite{zhou2019objects,liu2020smoke,chen2020monopair,li2020rtm3d} adhering to this keypoint-based paradigm have a similar structural layout, which is composed of a common backbone network followed by multiple parallel output heads. This layout is limited in terms of equally treating different output modalities (2D center, 3D center, size, rotation, depth, etc.) without considering their estimation difficulty. We find that by sequentially associating the output features in a sorted order, the detector's performance is considerably improved.

On the other hand, in the context of autonomous driving, the labels of input data are unevenly distributed along the vertical dimension of 2D images. For instance, the objects located near the top edge of the image tend to have smaller 2D bounding boxes and larger depth values than those near the bottom edge. M3D-RPN~\cite{brazil2019m3d} has exploited this distribution by devising depth-aware convolutions. We take a different pathway by generating row-wise features (named as depth hints), which are explicitly supervised in a bin-wise manner and used to augment the features for depth estimation.

In this work, a single-stage keypoint-based monocular 3D object detection network is presented, which is featured by sequential \textbf{f}eature \textbf{a}ssociation and \textbf{d}epth hint augmentation (FADNet). Since we are not the first to propose the keypoint-based framework for monocular 3D object detection, a brief comparison between previous keypoint-based methods~\cite{zhou2019objects,liu2020smoke,chen2020monopair,li2020rtm3d} and ours is displayed in Table \ref{tab1}.
\begin{table*}[htbp]
	\centering
	\caption{Comparison with other keypoint-based methods. The `Additional Branches' are defined with respect to the output branches of CenterNet~\cite{zhou2019objects}.}
	\label{tab1}
	\resizebox{0.85\textwidth}{!}{%
		\begin{tabular}{cccccc}
			\toprule
			\multirow{2}{*}{}&\multicolumn{3}{c}{Keypoint Estimation}&
			Sequential Feature&\multirow{2}{*}{Additional Branches}\\
			
			&2D center&3D centroids& 3D vertices & Association & \\
			\midrule
			CenterNet~\cite{zhou2019objects}
			&\checkmark&&&&n/a \\
			
			MonoPair~\cite{chen2020monopair}
			& & \checkmark & &&pairwise constraint \\
			
			SMOKE~\cite{liu2020smoke}
			& & \checkmark & &&None \\
			
			RMT3D~\cite{li2020rtm3d}
			& \checkmark & \checkmark & \checkmark &&vertices offset \\
			
			Ours~(FADNet)
			& \checkmark & \checkmark & &\checkmark&depth hint \\
			\bottomrule	
		\end{tabular}%
	}
\end{table*}
The first initiative of this work is the strategy of output grouping and sequential feature association. To be specific, the network output modalities are heuristically divided into several groups according to their estimation difficulty levels. The output modalities in different groups are estimated by output features from different timesteps of a convolutional gated recurrent unit (convGRU). The output features in the convGRU are sequentially associated by the propagation of hidden states. This has the advantage that the estimation difficulty of outputs from later timesteps is alleviated by the enlightenment of hidden states from previous timesteps. By experiment, it is found that such a design also contributes to promoting the consistency between 2D and 3D estimations, which is attributed to the associative structure in the proposed method.

The second initiative of this work is the strategy of depth hint augmentation. Specifically, a dedicated depth hint module is designed to learn a depth pattern from the input image, which serves as an informative hint for subsequent depth estimation. Due to the coherence of objects' depth values and 2D vertical locations, this learned depth pattern is instantiated by a row-wise feature called depth hint generated by a dedicated depth hint module. The depth hint is then used to augment the output features for predicting objects' depth, leading to enhancement in network performance.

To evaluate the contributions of this work, experiments are conducted on KITTI~\cite{geiger2012we}, which is the most prestigious benchmark for autonomous driving. Experimental results indicate that the presented FADNet is superior to the baseline and performs competitively in comparison with state-of-the-art monocular 3D object detection methods while maintaining a decent running speed. The contributions of the two initiatives in this work are further corroborated through ablation study. The code is available at \href{https://github.com/gtzly/FADNet}{\textit{https://github.com/gtzly/FADNet}}.

The main contributions in this work are summarized as follows:
\begin{itemize}
	\item A single-stage keypoint-based network, FADNet, is presented for monocular 3D object detection with autonomous driving as the target application.
	\item The strategy of output grouping and sequential feature association is proposed.
	\item A dedicated depth hint module is designed to generate depth hints, by which the convolutional features are augmented for better depth estimation.
	\item The presented FADNet performs competitively with state-of-the-art methods, providing a good trade-off between the detection accuracy and running speed.
\end{itemize}

\begin{figure*}[htbp]
	{\centering
		\begin{overpic}[width=0.85\textwidth]{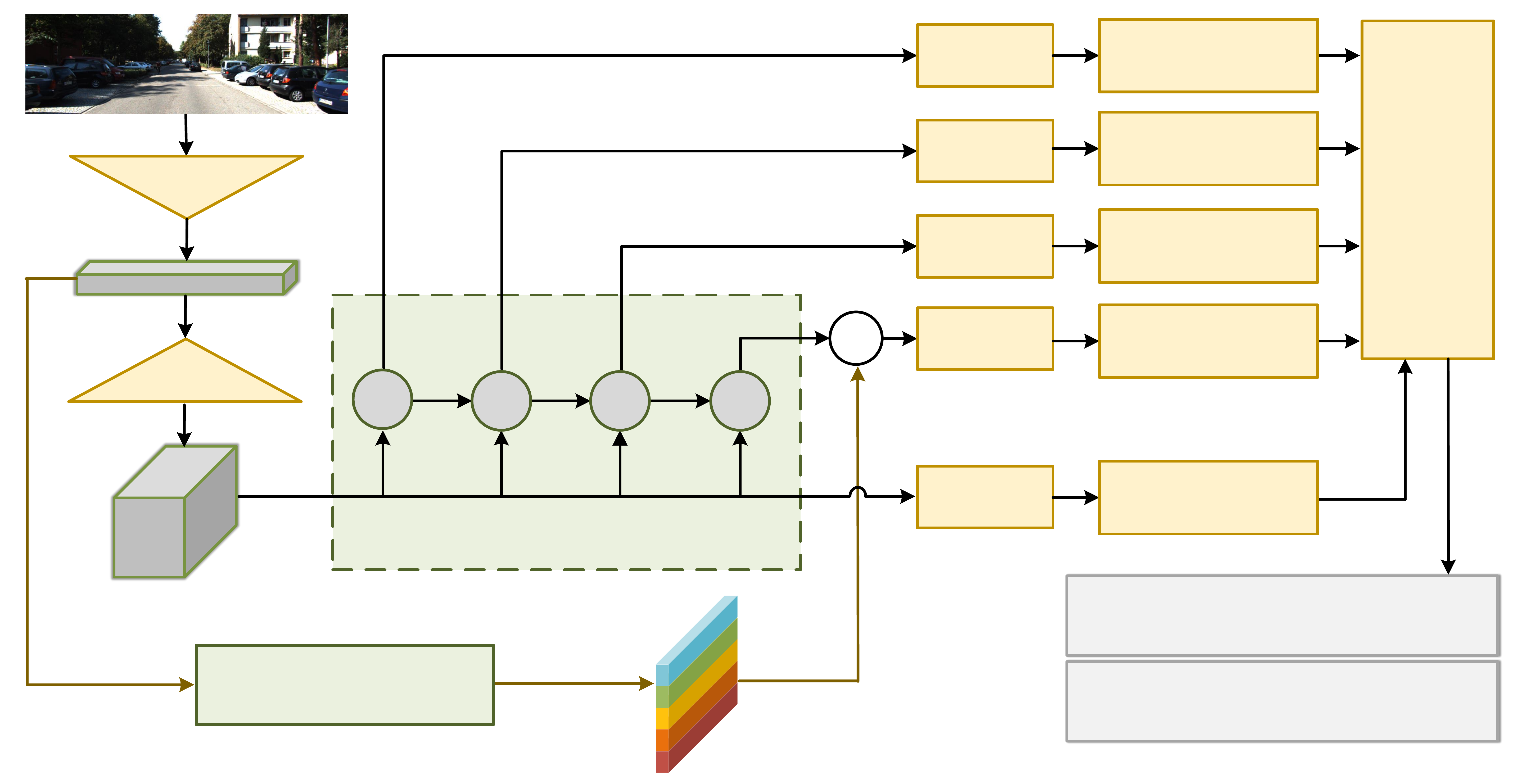}
			\put(10.2,39){\textcolor{black}{ {\footnotesize ${\rm DLA}$}}}
			\put(10.7,26.1){\textcolor{black}{{\footnotesize ${\rm UP}$}}}
			\put(14.4,5.8){\textcolor{black}{{\footnotesize ${\rm Depth \; Hint \; Module}$}}}
			
			\put(36,15){\textcolor{black}{{\small ${\rm G^{*}: convGRU}$}}}
			\put(55.7,28.6){\textcolor{black}{{\small ${\rm C}$}}}
			\put(47.8,24.5){\textcolor{black}{{\small ${\rm G^{*}}$}}}
			\put(40,24.5){\textcolor{black}{{\small ${\rm G^{*}}$}}}
			\put(31.7,24.5){\textcolor{black}{{\small ${\rm G^{*}}$}}}
			\put(24.2,24.5){\textcolor{black}{{\small ${\rm G^{*}}$}}}
			
			\put(61.1,47.3){\textcolor{black}{{\footnotesize ${\rm Reg \; Head}$}}}
			\put(73.2,47.3){\textcolor{black}{{\footnotesize $\Delta u_{2d},\! \Delta v_{2d},\! w, \! h$}}}
			
			\put(61.1,41.1){\textcolor{black}{{\footnotesize ${\rm Reg \; Head}$}}}
			\put(74.6,41.9){\textcolor{black}{{\footnotesize $\Delta H,\! \Delta W,\! \Delta L,$}}}
			\put(76.1,40.4){\textcolor{black}{{\footnotesize ${\rm cos}\alpha,\! {\rm sin}\alpha$}}}
			
			\put(61.1,35.1){\textcolor{black}{{\footnotesize ${\rm Reg \; Head}$}}}
			\put(75.3,35.1){\textcolor{black}{{\footnotesize $\Delta u_{3d},\Delta v_{3d}$}}}
			
			\put(61.1,28.6){\textcolor{black}{{\footnotesize ${\rm Reg \; Head}$}}}
			\put(79.2,28.4){\textcolor{black}{{\footnotesize $\hat{d}$}}}
			
			\put(61.1,18.3){\textcolor{black}{{\footnotesize ${\rm KP \; Head}$}}}
			\put(76.9,18.4){\textcolor{black}{{\footnotesize $u_{3d}^K,v_{3d}^K$}}}
			
			\put(91.3,39.2){\textcolor{black}{{\footnotesize ${\rm Output}$}}}
			\put(90.5,36.9){\textcolor{black}{{\footnotesize ${\rm Decoding}$}}}
			
			\put(76,10.3){\textcolor{black}{{\footnotesize ${\rm 2D \rightarrow }(u_{2d},v_{2d},w,h)$}}}
			\put(71.8,5){\textcolor{black}{{\footnotesize ${\rm 3D \rightarrow }(x_{3d},y_{3d},z_{3d},H,W,L,\theta)$}}}
			
		\end{overpic}
		\caption{Overall network architecture. `DLA' and `UP' stand for DLA-34 and the upsampling network. `C' denotes concatenation. `Reg' and `KP' are short for `Regression' and `Keypoint'.}
		\label{architecture}
	}
\end{figure*}
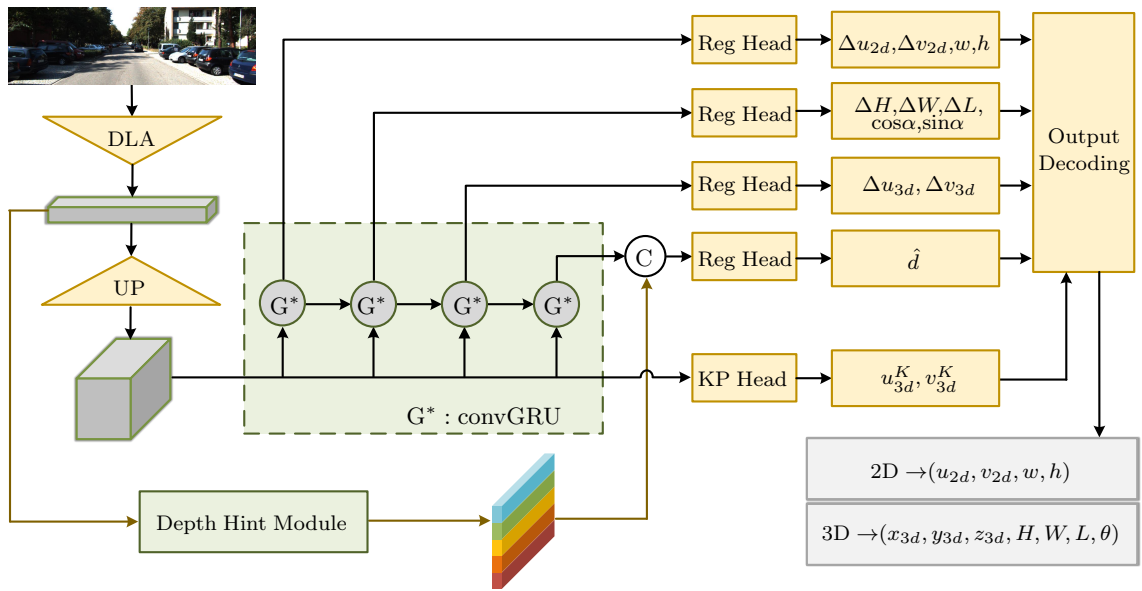

\section{Related Work}
In this section, 2D object detection methods are briefly reviewed in Subsection~\ref{2D_object_detection}. Then monocular 3D detection methods are reviewed from two aspects according to whether they are region proposal-based methods (Subsection~\ref{region_proposal_based}) or keypoint-based methods (Subsection~\ref{keypoint_based}).

\subsection{2D Object Detection}
\label{2D_object_detection}
2D object detection methods are conventionally categorized into two-stage methods~\cite{girshick2014rich,girshick2015fast,ren2015faster,he2015spatial,lin2017feature} and one-stage methods~\cite{liu2016ssd,lin2017focal,redmon2016you,redmon2017yolo9000,redmon2018yolov3,zhou2019objects}. The most representative two-stage detector is Faster R-CNN~\cite{ren2015faster}, which is the first to devise the region proposal network.
FPN~\cite{lin2017feature} has integrated a feature pyramid network into Faster R-CNN to address the issue of multi-scale detection. RefineNet~\cite{rajaram2016refinenet} has proposed a framework that iteratively refines the region proposals generated by a Faster R-CNN. Despite of being relatively complex and time-consuming, two-stage detectors generally achieve higher precision and recall rates in comparison with the one-stage counterparts.

To narrow down the performance gap between one-stage and two-stage approaches, RetinaNet~\cite{lin2017focal} has proposed to mitigate the imbalance between the number of positive and negative samples by using a Focal Loss. YOLOv3~\cite{redmon2018yolov3} has presented an efficient Darknet-53 backbone for feature extraction. Combined with multiple output heads and predefined anchors, YOLOv3 achieves a decent trade-off between running speed and detection accuracy. More recently, YOLOv4~\cite{bochkovskiy2020yolov4} has further incorporated multiple technical tricks into YOLOv3, yielding a network that achieves better performance and is easier to be trained.

More recently, some anchor-free detectors~\cite{law2018cornernet,tian2019fcos,zhou2019bottom,yang2019reppoints,lu2019grid} have been proposed. CornerNet~\cite{law2018cornernet} has phrased the 2D detection tasks as the keypoint detection of top-left and bottom-right corners.
FCOS~\cite{tian2019fcos} has presented a per-pixel dense prediction approach, which regresses the distances from a pixel to four edges of the potential bounding boxes it may reside in.
ExtremeNet~\cite{zhou2019bottom} has proposed to estimate the center point together with four extreme points of 2D bounding boxes. A brute force strategy is leveraged to group all quadruples of the estimated extreme points.
This line of works are free from the manually predefined anchors, yet still achieve promising results.

\subsection{Region Proposal-based Monocular 3D Detection}
\label{region_proposal_based}
Similar to two-stage 2D object detection methods~\cite{girshick2014rich,girshick2015fast,ren2015faster,he2015spatial}, many researchers~\cite{chen2016monocular,xiang2017subcategory,mousavian20173d,chabot2017deep,naiden2019shift,bao2019monofenet,he2019mono3d++,hu2019joint,manhardt2019roi,ku2019monocular} have sought to predict 3D information by leveraging the power of region proposals to reduce the search scope.
Deep3DBox~\cite{mousavian20173d} and Shift R-CNN~\cite{naiden2019shift} have proposed to estimate the 3D object locations by solving a linear system which is established through geometric constraints between 2D and 3D bounding boxes.
Deep MANTA~\cite{chabot2017deep} has used a coarse-to-fine region proposal network to detect 2D vehicular parts. The task of 3D object detection is then formulated as a PnP problem and solved by a 2D/3D matching algorithm~\cite{lepetit2009epnp}.
MonoGRNet~\cite{qin2019monogrnet} has firstly designed an instance-level depth estimation network. The merits of both early features and deep features are exploited to obtain a refined depth estimation.
MonoFENet~\cite{bao2019monofenet}, ROI-10D~\cite{manhardt2019roi} and Mono3D++~\cite{he2019mono3d++} have proposed to leverage a monocular disparity/depth estimation network to provide extra depth priors.
MonoDIS~\cite{simonelli2019disentangling} has introduced the methodology of loss disentanglement to refine the optimization dynamics.
M3D-RPN~\cite{brazil2019m3d} has proposed to ease the difficulty in monocular 3D detection by a series of 2D and 3D predefined anchors. Notably, albeit being region proposal-based methods, the detection pipeline of M3D-RPN~\cite{brazil2019m3d} is designed in a one-stage manner.

\subsection{Keypoint-based Monocular 3D Detection}
\label{keypoint_based}
Another line of works~\cite{zhou2019objects,liu2020smoke,chen2020monopair,li2020rtm3d} have followed the keypoint-based scheme, which is firstly proposed by the pioneering work CenterNet~\cite{zhou2019objects} from Zhou et al. CenterNet phrases 2D/3D object detection as predicting the 2D bounding box centers via a keypoint branch and other geometric properties via several class-agnostic regression branches. Because CenterNet is originally oriented towards 2D detection tasks, it has not deep dived into its 3D extension and achieves relatively poor results on the KITTI\cite{geiger2012we} benchmark.

Later keypoint-based approaches~\cite{liu2020smoke,chen2020monopair,li2020rtm3d} have generally retained CenterNet's architecture yet have made effective improvements to adapt to 3D detection tasks.
SMOKE~\cite{liu2020smoke} has proposed to predict projected 3D centroids instead of 2D bounding box centers. The loss disentanglement method by~\cite{simonelli2019disentangling} has been extended to a multi-step form to facilitate the training stage.
MonoPair~\cite{chen2020monopair} has designed an additional pairwise constraint branch to exploit the mutual spatial relationship between vehicles for further post-optimization. The weights for error minimization are provided by the predicted aleatoric uncertainty.
RMT3D~\cite{li2020rtm3d} has presented a network that predicts the centers of 2D bounding boxes together with another 9 keypoints per object, including the projected 3D centroid and 8 projected vertices of the 3D bounding box.

Our work is inspired by keypoint-based 3D detection frameworks, which have a similar architecture in view of the parallel layout of multiple output heads based on a common backbone network for feature extraction. However, previous works are limited due to having an implicit association across different output modalities (2D center, 3D center, size, rotation, depth, etc.) and ignoring the difference in estimation difficulty between them. By contrast, our method tackles both drawbacks through the strategy of output grouping and sequential feature association. Besides, we present a depth hint augmentation method to better exploit the depth patterns of input images.

\section{Methodology}
In this section, the network architecture is firstly described in Subsection~\ref{network_architecture}. Then, two initiatives of our method, i.e. sequential feature association and depth hint augmentation, are explained in Subsection~\ref{feature_association} and \ref{depth_hint_augmentation}. Finally, the design of loss functions is introduced in Subsection~\ref{loss_functions}.

\subsection{Network Architecture}
\label{network_architecture}
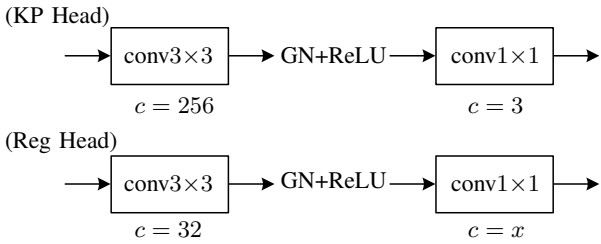
\begin{figure}[htbp]
	{\centering
		\begin{overpic}[width=0.47\textwidth]{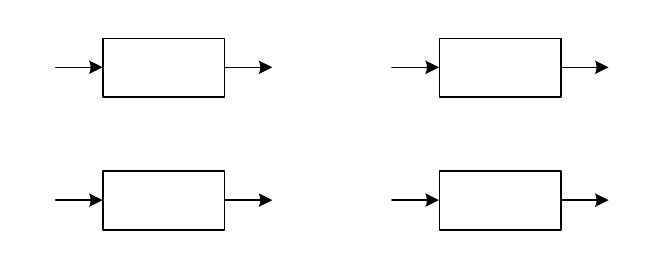}
			\put(17.6,29){\textcolor{black}{{\small conv$3\PLH3$}}}		
			\put(68.6,29){\textcolor{black}{{\small conv$1\PLH1$}}}	
			\put(19.2,21.3){\textcolor{black}{{\small $c=256$}}}	
			\put(71.2,21.3){\textcolor{black}{{\small $c=3$}}}	
			\put(42,29){\textcolor{black}{{\small GN+ReLU}}}
			\put(-1,35.3){\textcolor{black}{{\small (KP Head)}}}	
			
			\put(17.6,9){\textcolor{black}{{\small conv$3\PLH3$}}}		
			\put(68.6,9){\textcolor{black}{{\small conv$1\PLH1$}}}
			\put(19.2,1.8){\textcolor{black}{{\small $c=32$}}}	
			\put(71.2,1.8){\textcolor{black}{{\small $c=x$}}}
			\put(42,9.6){\textcolor{black}{{\small GN+ReLU}}}
			\put(-1,15.8){\textcolor{black}{{\small (Reg Head)}}}

		\end{overpic}
		\caption{Structures of the keypoint head and regression heads. `GN' and `ReLU' stand for Group Normalization~\cite{wu2018group} and Rectified Linear Unit, respectively. `c' denotes the number of output channels and `x' depends on the output grouping strategy (Subsection~\ref{feature_association}).}
		\label{head}
	}
\end{figure}

\textbf{Overview}
A single-stage, keypoint-based network is presented for monocular 3D detection in autonomous driving scenarios.
The overall architecture is shown in Fig. \ref{architecture}. Given an input image with a resolution of $ W \times H$, a backbone network with an output stride of $S$ is employed for feature extraction, resulting in an output feature with a size of $ W/S \times H/S$. A keypoint head and multiple regression heads are utilized to output 2D and 3D information on the basis of the extracted feature. The structures of head networks are shown in Fig.~\ref{head}.

Different from previous works, the convolutional feature extracted by the backbone network is further sent into a convolutional gated recurrent unit (convGRU). Besides, in order to exploit the depth patterns of input images, depth hints are extracted by a depth hint module and then used to augment the output feature for depth estimation. Details of these two initiatives will be further elaborated on in later sections.

\textbf{Backbone}
A deep layer aggregation network DLA-34 \cite{yu2018deep} combined with an upsampling network is employed as the backbone network in our method. The DLA-34 network has an output stride of 32, producing an intermediate feature of $ W/32 \times H/32 \times 512$. Inspired by \cite{zhou2019objects}, we use the same upsampling network where transposed convolutions are utilized to increase the feature resolution and deformable convolutional layers \cite{zhu2019deformable} are added ahead of upsampling layers. The final output feature has a shape of  $ W/4 \times H/4 \times 64$.

\textbf{Keypoint Branch}
A $ W/4 \times H/4 \times C$ sized heatmap is predicted by the keypoint head network, where $C$ is the number of object categories ($C=3$ for KITTI\cite{geiger2012we}). The value of each element in the heatmap indicates how possible the projected 3D centroid of a certain object category exists at the same location. By further calculating the local maxima and filtering via a threshold, we can obtain a preliminary estimation of the projected 3D centroids, denoted by $K_{3d}=[u_{3d}^K,v_{3d}^K]^T$. Since the output of keypoint branch has a stride of 4, discretization errors exist between the estimated keypoints $K_{3d} $ and the accurate locations of projected 3D centroids $C_{3d} $. Such errors will be further estimated by the 3D regression branch in later description.

\begin{figure}[htbp]
	{\centering
		\begin{overpic}[width=0.46\textwidth]{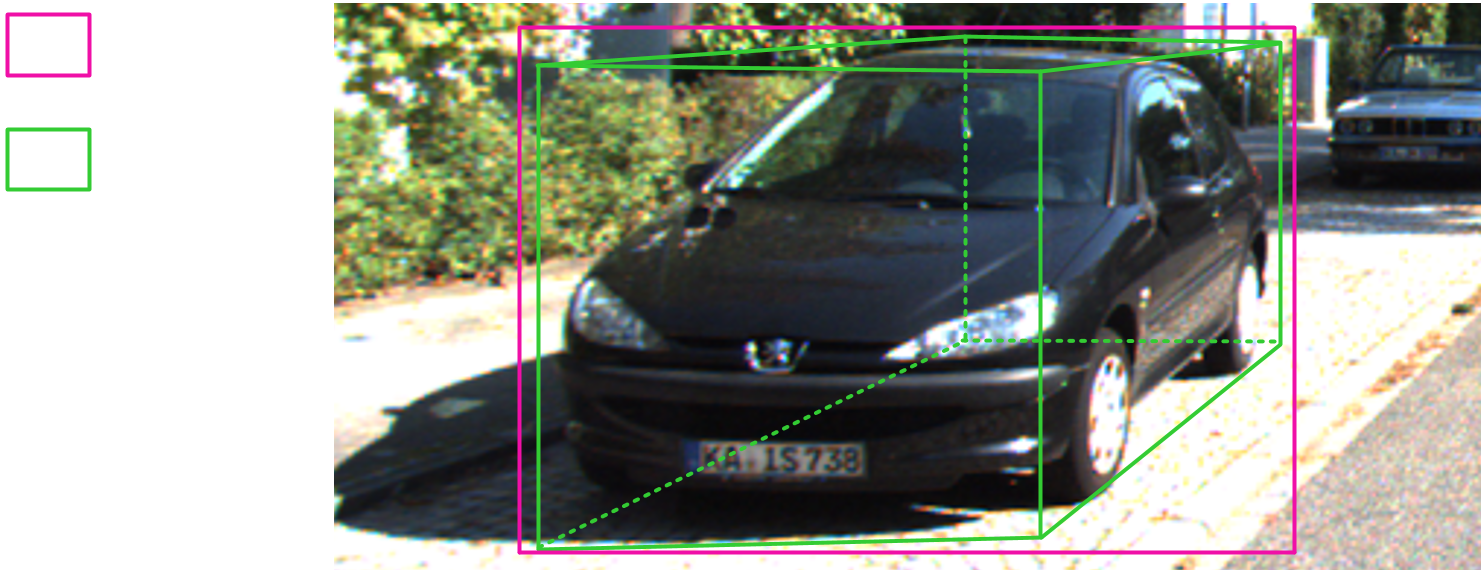}
			\put(8,34.6){\textcolor{black}{{\small 2D box}}}
			\put(8,26.8){\textcolor{black}{{\small 3D box}}}		

		\end{overpic}
		\caption{Illustration of cases when labeled 2D bounding boxes do not wrap tightly around the projection of 3D bounding boxes, best viewed in color.}
		\label{box_fitting}
	}
\end{figure}

\textbf{2D Regression Branch}
To obtain a better understanding of 2D semantic properties, it is desirable to learn 2D bounding boxes' centers $C_{2d}$ and 2D dimensions ($w, h$). To this end, the 2D regression branch outputs a tuple of 4 elements ($\Delta u_{2d}, \Delta v_{2d}, w, h$), where ($\Delta u_{2d}, \Delta v_{2d}$) denote the class-agnostic offsets between $C_{2d}$ and $K_{3d}$. The locations of 2D centers are thus decoded as
\begin{equation}
C_{2d}=
\begin{bmatrix}
u_{2d} \\
v_{2d}
\end{bmatrix}=
\begin{bmatrix}
u_{3d}^K+\Delta u_{2d}\\
v_{3d}^K+\Delta v_{2d}
\end{bmatrix}
\end{equation}

When setting up the ground truth labels for training the 2D head, it is found that the labeled 2D bounding boxes do not always wrap tightly around the projection of 3D bounding boxes (Fig.~\ref{box_fitting}). This phenomenon contradicts with our designing purpose in that inconsistent information would flow from 2D output features to 3D output features via sequential feature association (Subsection~\ref{feature_association}). Thus, instead of training with the raw labels provided by KITTI dataset, we remake the 2D labels with 2D boxes that tightly enclose the extreme points (left, top, right, bottom) of the projected 3D bounding boxes.

\vspace{-0.2cm}
\begin{figure}[h]
	{\centering
		\begin{overpic}[width=0.33\textwidth]{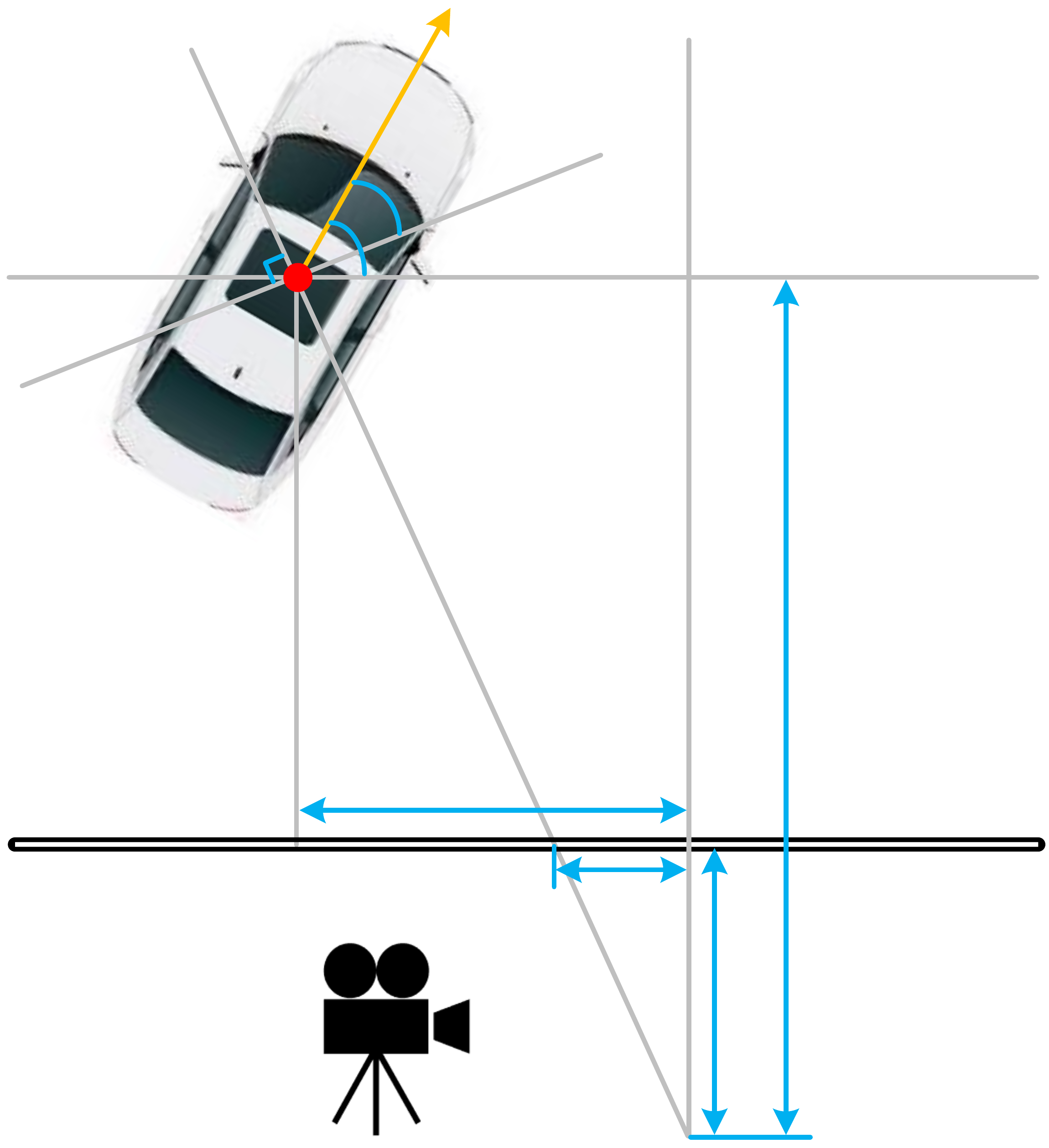}
			\put(35.9,82.6){\textcolor{cyan}{\boldmath {\footnotesize $\alpha$}}}
			\put(31.3,78.9){\textcolor{cyan}{\boldmath {\footnotesize $\theta$}}}
			\put(40,31){\textcolor{cyan}{\boldmath {\footnotesize $x_{3d}$}}}
			\put(70,40){\textcolor{cyan}{\boldmath {\footnotesize $d\,(z_{3d})$}}}
			\put(62.5,12){\textcolor{cyan}{\boldmath {\footnotesize $f_x$}}}
			\put(54,1){\textcolor{cyan}{\boldmath {\footnotesize $O$}}}
			\put(45,19){\textcolor{cyan}{\boldmath {\footnotesize $u_{3d}\scalebox{0.55}[0.8]{\( - \)}u_0$}}}
			\put(73,28){\textcolor{black}{\text {\footnotesize $ \rm{ image \;plane } $ }}}
		\end{overpic}
		\caption{Illustration of the geometric relationships in 3D output decoding }
		\label{decoding}
	}
\end{figure}
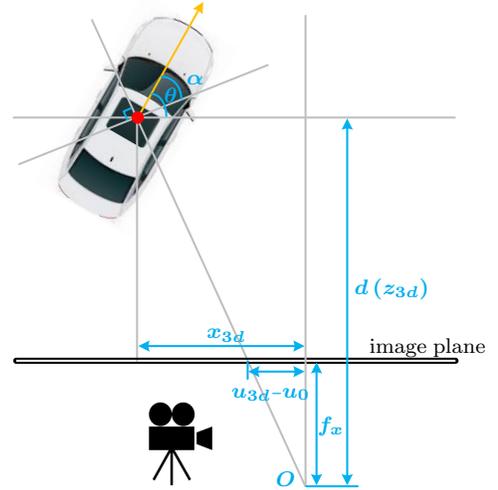
\vspace{-0.2cm}
\textbf{3D Regression Branch}
Similar to \cite{zhou2019objects,chen2020monopair,liu2020smoke}, our 3D regression branch estimates the following variables.
\begin{itemize}
	\item The discretization error $[\Delta u_{3d}, \Delta v_{3d}]^T$ between the accurate location of the projected 3D centroid $C_{3d}$ and its keypoint estimation $K_{3d}$, i.e. $[\Delta u_{3d}, \Delta v_{3d}]^T=[u_{3d}-v_{3d}^K,v_{3d}-v_{3d}^K]^T$.
	\item The cosine and sine values of the viewing angle $[{\rm cos}\alpha,{\rm sin}\alpha ]^T$. The difference between the viewing angle $\alpha$ and the yaw angle $\theta$ is illustrated in Fig.~\ref{decoding}.
	\item The exponential offset $[\Delta H , \Delta W ,\Delta L]^T $ between 3D dimensions $[H,W,L]^T$ and the class-specific dimension template $[\bar{H},\bar{W},\bar{L}]^T$, which is calculated by averaging over the training dataset. That is, $[H,W,L]^T=[\bar{H}{\rm exp}(\Delta H),\bar{W}{\rm exp}(\Delta W),\bar{L}{\rm exp}(\Delta L)]^T$.
	\item The encoded depth $ \widetilde{d} $ of the projected 3D centroid following \cite{eigen2014depth}. The real depth value $ d $ is calculated as $d=1/{\rm \sigma}(\widetilde{d})-1$ , with $\sigma$ denoting the sigmoid function.
\end{itemize}

Given aforementioned estimations and the camera's intrinsic matrix
\begin{equation}
K=\begin{bmatrix}
f_x&0&u_0 \\
0&f_y&v_0 \\
0&0&1
\end{bmatrix},
\end{equation}
where $f_x$ and $f_y$ are focal lengths, ($u_0, v_0$) is the coordinate of the principal point. The 3D location can then be decoded using the projection equation $z_{3d}[u_{3d},v_{3d},1]^T=K[x_{3d},y_{3d},z_{3d}]^T$. Specifically, we have

\begin{equation}
C_{3d}=
\begin{bmatrix}
x_{3d} \\
y_{3d} \\
z_{3d}
\end{bmatrix}=
\begin{bmatrix}
z_{3d}(u_{3d}^K+\Delta u_{3d}-u_0)/f_x\\
z_{3d}(v_{3d}^K+\Delta v_{3d}-v_0)/f_y\\
1/{\rm \sigma}(\hat{d})-1
\end{bmatrix}
\end{equation}

The yaw angle $\theta$ in the camera coordinate is decoded as (Fig.~\ref{decoding})
\begin{equation}
\theta=\alpha+{\rm arctan}(x_{3d}/z_{3d})
\end{equation}

\subsection{Sequential Feature Association}
\label{feature_association}
\textbf{Output Grouping}
The outputs of regression heads are heuristically divided into four groups according to the estimation difficulty. The first group contains the 2D properties ($\Delta u_{2d}, \Delta v_{2d}, w, h$), which are the most intuitive and easiest to estimate from the 2D image. The second group contains the 3D dimension offsets ($\Delta H, \Delta W, \Delta L$) and the cosine and sine values of the viewing angle $({\rm cos}\alpha,{\rm sin}\alpha)$, which are a bit more abstract yet still can be deduced from 2D appearances. The third group contains the discretization errors of projected 3D centroids ($ \Delta u_{3d}, \Delta v_{3d} $), which directly affect the final localization accuracy. The last group contains the encoded depths $\hat{d}$, which requires a comprehensive understanding of the objective appearances and surrounding contexts to achieve an accurate estimation.

\textbf{Association via ConvGRU}
The strategies of output grouping and sequential feature association can be naturally implemented by assigning different output groups to different timesteps of a recurrent neural network (Fig.~\ref{architecture}). In practice, we use a single-layer GRU network which is adapted into a convolutional version to maintain the spatial structure of features, namely the multiplications in a normal GRU are replaced with convolutions. All the convolutions in convGRU are set as stride=$1$, kernel size=$3 \time 3$, and output channels=$64$.

Mathematically, given an input image $I$, the output of the regression branch at timestep $t\;(t=1,2,3,4)$ can be represented as
\begin{equation}
y_t=\Phi_t(B(I),h_{t-1}),
\end{equation}
where $B(\cdot)$ denotes the backbone network. $\Phi_t(\cdot)$ denotes the head network for timestep $t$ (or equally, the $t^{\rm th}$ output group). $h_{t-1}$ stands for the hidden state in the convGRU at timestep $t-1$. By this formulation, it can be seen that the network outputs at later timesteps have access to the hidden states flowing from former timesteps. Since the output groups are arranged according to the estimation difficulty, the harder estimations are benefited from the prior guidance of easier ones. For example, being aware that the object has a small 2D dimension, the network is more likely to output a large depth estimation.

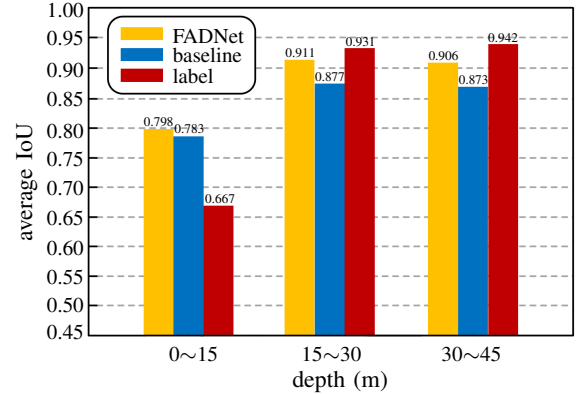
\begin{figure}[htbp]
	{\centering
		\begin{overpic}[width=0.45\textwidth]{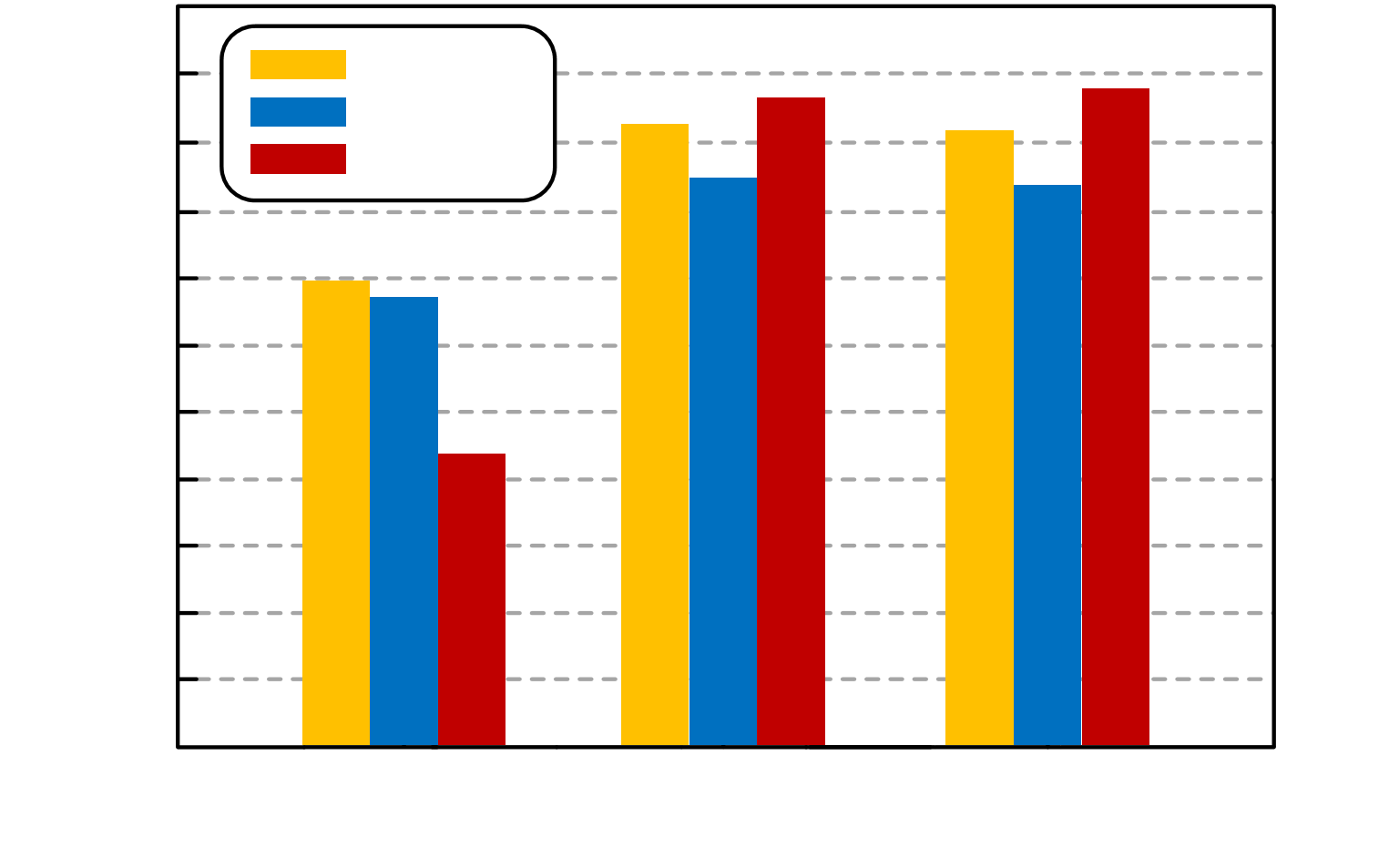}
			\put(1,24){\textcolor{black}{\boldmath {\small \rotatebox{90}{average IoU}}}}
			
			\put(6,8){\textcolor{black}{ {\footnotesize 0.45}}}
			\put(6,12.5){\textcolor{black}{ {\footnotesize 0.50}}}
			\put(6,17){\textcolor{black}{ {\footnotesize 0.55}}}
			\put(6,21.7){\textcolor{black}{ {\footnotesize 0.60}}}
			\put(6,26.6){\textcolor{black}{ {\footnotesize 0.65}}}
			\put(6,31.6){\textcolor{black}{ {\footnotesize 0.70}}}
			\put(6,36.2){\textcolor{black}{ {\footnotesize 0.75}}}
			\put(6,40.9){\textcolor{black}{ {\footnotesize 0.80}}}
			\put(6,46){\textcolor{black}{ {\footnotesize 0.85}}}
			\put(6,50.8){\textcolor{black}{ {\footnotesize 0.90}}}
			\put(6,56){\textcolor{black}{ {\footnotesize 0.95}}}
			\put(6,60.5){\textcolor{black}{ {\footnotesize 1.00}}}

			\put(25.9,4.5){\textcolor{black}{ {\footnotesize 0$\sim$15}}}
			\put(47.5,4.5){\textcolor{black}{ {\footnotesize 15$\sim$30}}}
			\put(70.3,4.5){\textcolor{black}{ {\footnotesize 30$\sim$45}}}
			\put(46,0.3){\textcolor{black}{ {\small depth (m)}}}
			
			\put(26.5,56.4){\textcolor{black}{ {\footnotesize FADNet}}}
			\put(26.5,53.2){\textcolor{black}{ {\footnotesize baseline}}}
			\put(26.5,49.76){\textcolor{black}{ {\footnotesize label}}}
			
			\put(21.8,42.8){\textcolor{black}{ {\tiny 0.798}}}
			\put(26.8,41.6){\textcolor{black}{ {\tiny 0.783}}}
			\put(31.8,30.2){\textcolor{black}{ {\tiny 0.667}}}
			
			\put(44.9,54.1){\textcolor{black}{ {\tiny 0.911}}}
			\put(49.7,50.1){\textcolor{black}{ {\tiny 0.877}}}
			\put(54.7,55.9){\textcolor{black}{ {\tiny 0.931}}}

			\put(68.2,53.5){\textcolor{black}{ {\tiny 0.906}}}
			\put(73.1,49.6){\textcolor{black}{ {\tiny 0.873}}}
			\put(77.9,56.5){\textcolor{black}{ {\tiny 0.942}}}
			
		\end{overpic}
		\caption{Statistics of 2D-3D consistency with respect to objects' depths for the `car' class of the validation set of KITTI. The `label' denotes the ground truth labels provided by KITTI. Noting that the average IoU of `label' is lower than the others for objects with depths within 0$\sim$15m. This is because objects near the camera are often truncated, while the ground truth labels for 2D bounding boxes are delimited by the image boundary.}
		\label{consistency}
	}
\end{figure}

\textbf{2D-3D Consistency}
By experiments, it is found that the associative structure of FADNet contributes to the consistency between 2D and 3D estimations. For illustration, we calculate the average intersection over union (IoU) between the estimated 2D bounding boxes and the bounding boxes obtained by projecting the estimated 3D boxes onto the image plane. A baseline is designed for comparison by removing the sequential feature association from FADNet (i.e. without the convGRU). The resulted statistics with respect to objects' depths are shown in Fig.~\ref{consistency}. It can be observed that the 3D estimations of FADNet better coincide with the 2D estimations than the baseline, which implicitly indicates that the 3D estimations have been guided by the 2D estimations in FADNet.

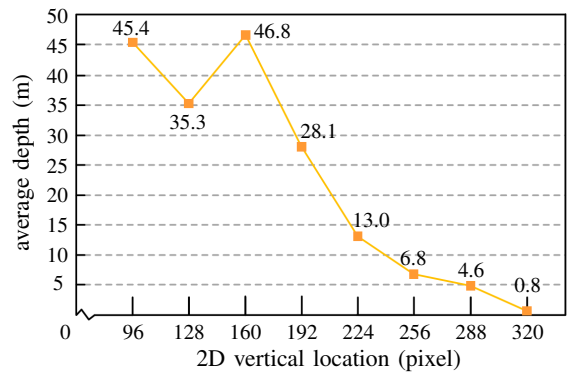
\begin{figure}[h]
	{\centering
		\begin{overpic}[width=0.46\textwidth]{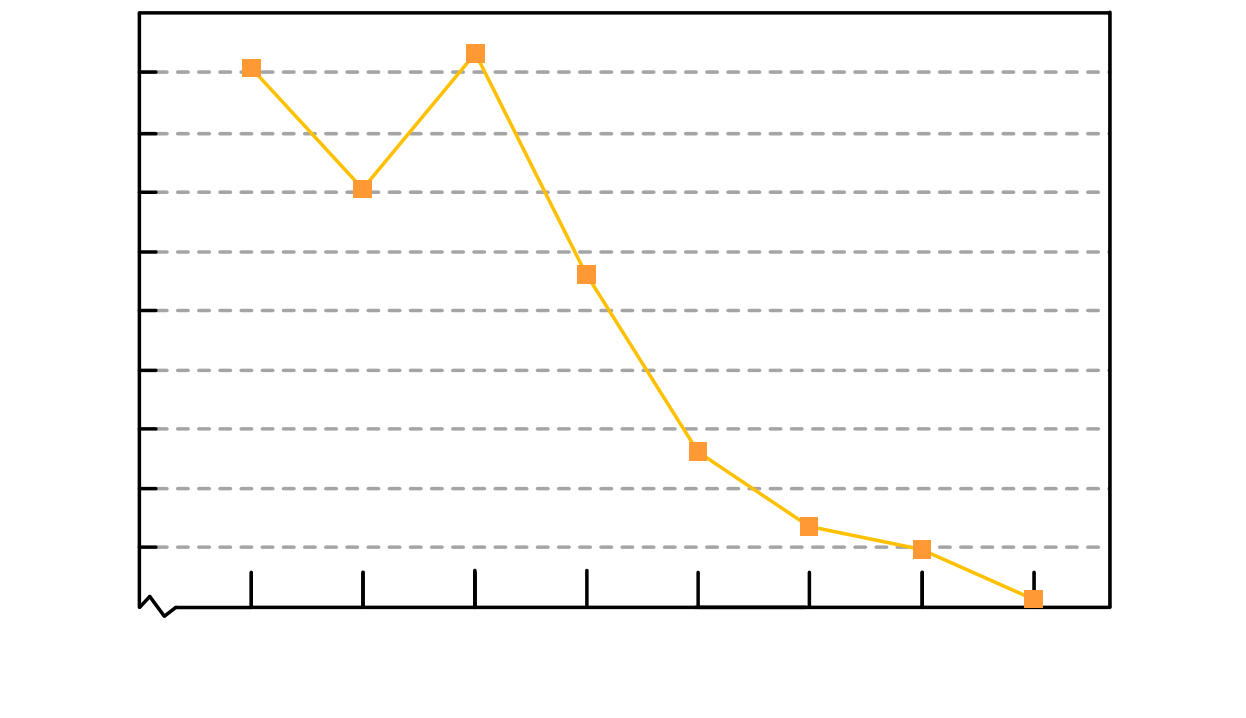}
			\put(17,53.5){\textcolor{black}{{\footnotesize 45.4}}}
			\put(26,38.5){\textcolor{black}{{\footnotesize 35.3}}}
			\put(39.5,53){\textcolor{black}{{\footnotesize 46.8}}}
			\put(46.9,37){\textcolor{black}{{\footnotesize 28.1}}}
			\put(55.2,22.8){\textcolor{black}{{\footnotesize 13.0}}}
			\put(62.8,16.7){\textcolor{black}{{\footnotesize 6.8}}}
			\put(72,14.8){\textcolor{black}{{\footnotesize 4.6}}}
			\put(81,12.4){\textcolor{black}{{\footnotesize 0.8}}}

			\put(8.6,5){\textcolor{black}{{\footnotesize 0}}}
			\put(18.6,5){\textcolor{black}{{\footnotesize 96}}}
			\put(26.6,5){\textcolor{black}{{\footnotesize 128}}}
			\put(35.6,5){\textcolor{black}{{\footnotesize 160}}}
			\put(44.6,5){\textcolor{black}{{\footnotesize 192}}}
			\put(53.6,5){\textcolor{black}{{\footnotesize 224}}}
			\put(62.6,5){\textcolor{black}{{\footnotesize 256}}}
			\put(71.6,5){\textcolor{black}{{\footnotesize 288}}}
			\put(80.6,5){\textcolor{black}{{\footnotesize 320}}}

			\put(7.5,12.7){\textcolor{black}{{\footnotesize 5}}}
			\put(6.5,17.3){\textcolor{black}{{\footnotesize 10}}}
			\put(6.5,22.1){\textcolor{black}{{\footnotesize 15}}}
			\put(6.5,26.8){\textcolor{black}{{\footnotesize 20}}}
			\put(6.5,31.5){\textcolor{black}{{\footnotesize 25}}}
			\put(6.5,36.2){\textcolor{black}{{\footnotesize 30}}}
			\put(6.5,41.1){\textcolor{black}{{\footnotesize 35}}}
			\put(6.5,45.8){\textcolor{black}{{\footnotesize 40}}}
			\put(6.5,50.6){\textcolor{black}{{\footnotesize 45}}}
			\put(6.5,55.3){\textcolor{black}{{\footnotesize 50}}}
			\put(30.3,0.7){\textcolor{black}{{\small 2D vertical location (pixel)}}}
			\put(1,19){\textcolor{black}{\boldmath {\small \rotatebox{90}{average depth (m)}}}}
		\end{overpic}
		\caption{Statistics of average depths with respect to 2D vertical locations in the image for the `car' class of the training set of KITTI.}
		\label{depth_vs_vertical}
	}
\end{figure}

\subsection{Depth Hint Augmentation}
\label{depth_hint_augmentation}
To better understand the depth patterns in autonomous driving scenarios, the statistics of average depths of cars at different 2D vertical locations are reported in Fig.~\ref{depth_vs_vertical}.
Two characteristics are observed from the statistics:

\romannumeral1)~The depths have a large variance, with values ranging from ${\rm <1}$m to ${\rm >45}$m.

\romannumeral2)~The depths are not uniformly distributed and have strong correlation with 2D vertical locations.

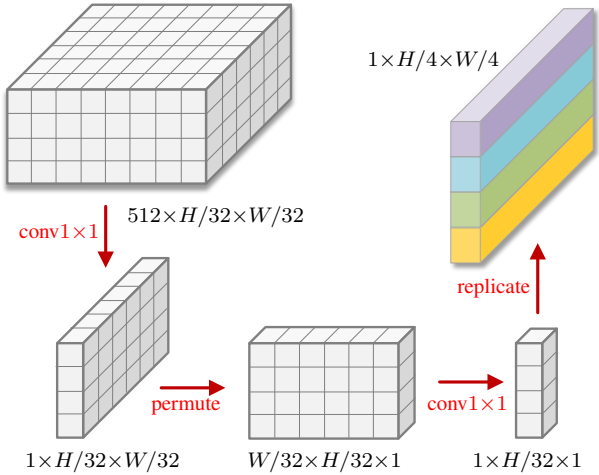
\begin{figure}[h]
	{\centering
		\begin{overpic}[width=0.44\textwidth]{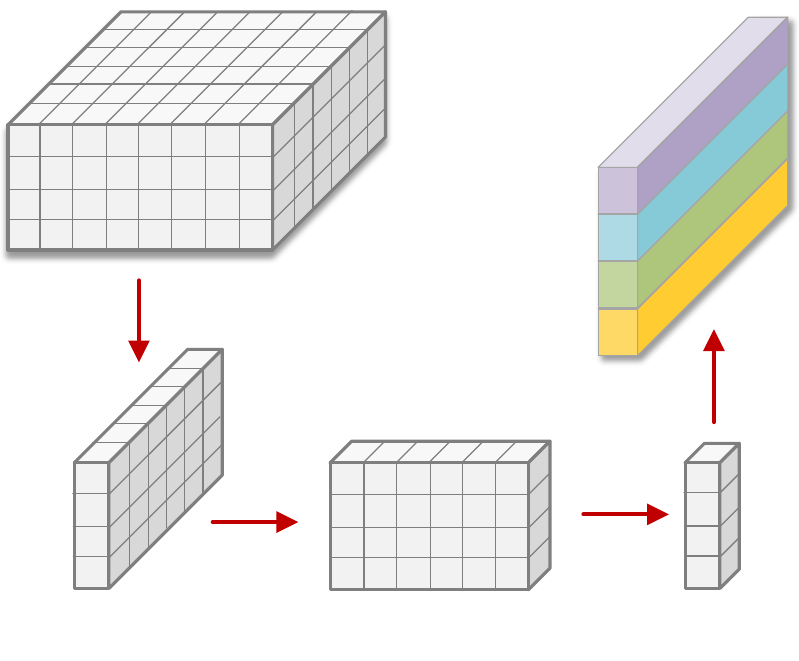}
			\put(21,45){\textcolor{black}{ {\footnotesize $512\PLH H/32\PLH W/32$}}}
			\put(4,4){\textcolor{black}{{\footnotesize $1\PLH H/32 \PLH W/32$}}}
			\put(41,4){\textcolor{black}{ {\footnotesize $W/32\PLH H/32 \PLH1$}}}
			\put(78.5,4){\textcolor{black}{ {\footnotesize $1\PLH H/32 \PLH1$}}}
			\put(61,71){\textcolor{black}{ {\footnotesize $1\PLH H/4\PLH W/4$}}}
			
			\put(3,42.2){\textcolor{red}{ {\footnotesize conv$1\PLH1$}}}	
			\put(25,13.5){\textcolor{red}{{\footnotesize permute}}}
			\put(71,13.5){\textcolor{red}{ {\footnotesize conv$1\PLH1$}}}
			\put(76,33){\textcolor{red}{{\footnotesize replicate}}}
				\vspace{-1cm}
		\end{overpic}
	\vspace{-0.4cm}
		\caption{Workflow of the depth hint module. $H$, $W$ correspond to the resolution of input image. The number of square blocks are not equivalent to the true sizes of features.}
		\label{depth_hint}
	}
\end{figure}

Therefore, it is favorable for the network to estimate objects' depths with the aid of some kind of row-wise references (or hints). Out of such motivation, a depth hint module is devised to generate such hints for depth estimation. Its workflow is illustrated in Fig.~\ref{depth_hint}. The feature fed into the depth hint module is output by the DLA-34 network without being upsampled (refer to Fig.~\ref{architecture}). By using convolutions with $1\times1$ sized kernels, the input feature is squeezed along all dimensions but the vertical one, resulting in a $H/32$ sized feature vector called depth hint vector. The depth hint vector is then replicated into a $1\times H/4\times W/4$ feature named as depth hint, which contains the estimated depth pattern along the vertical dimension. The features for depth estimation are augmented by being concatenated with these depth hints.

\begin{figure}[htbp]
	{\centering
		\begin{overpic}[width=0.5\textwidth]{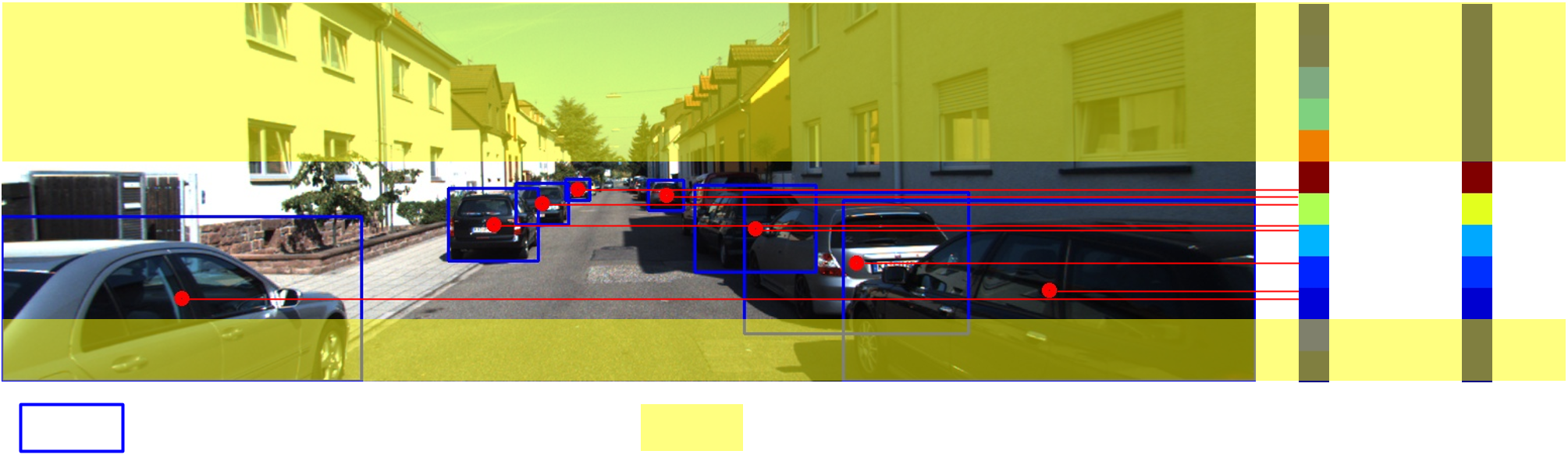}
			\put(10,1.1){\textcolor{black}{ {\small 2D bounding box}}}
			\put(50,1.1){\textcolor{black}{ {\small masked area}}}
			\put(81,1.1){\textcolor{black}{ {\small pred}}}
			\put(93,1.1){\textcolor{black}{ {\small gt}}}
			
			\put(85.2,27.7){\textcolor{black}{ {\tiny 0.06}}}
			\put(85.2,25.7){\textcolor{black}{ {\tiny 0.07}}}
			\put(85.2,23.7){\textcolor{black}{ {\tiny 10.58}}}
			\put(85.2,21.7){\textcolor{black}{ {\tiny 14.98}}}
			\put(85.2,19.7){\textcolor{black}{ {\tiny 49.91}}}
			\put(85.2,17.7){\textcolor{black}{ {\tiny 55.13}}}
			\put(85.2,15.7){\textcolor{black}{ {\tiny 29.55}}}
			\put(85.2,13.7){\textcolor{black}{ {\tiny 15.95}}}
			\put(85.2,11.7){\textcolor{black}{ {\tiny 8.06}}}
			\put(85.2,9.7){\textcolor{black}{ {\tiny 3.81}}}
			\put(85.2,7.7){\textcolor{black}{ {\tiny 3.67}}}
			\put(85.2,5.7){\textcolor{black}{ {\tiny 0.01}}}
			
			\put(95.5,27.7){\textcolor{black}{ {\tiny n/a}}}
			\put(95.5,25.7){\textcolor{black}{ {\tiny n/a}}}
			\put(95.5,23.7){\textcolor{black}{ {\tiny n/a}}}
			\put(95.5,21.7){\textcolor{black}{ {\tiny n/a}}}
			\put(95.5,19.7){\textcolor{black}{ {\tiny n/a}}}
			\put(95.5,17.7){\textcolor{black}{ {\tiny 55.04}}}
			\put(95.5,15.7){\textcolor{black}{ {\tiny 32.97}}}
			\put(95.5,13.7){\textcolor{black}{ {\tiny 16.01}}}
			\put(95.5,11.7){\textcolor{black}{ {\tiny 9.60}}}
			\put(95.5,9.7){\textcolor{black}{ {\tiny 4.35}}}
			\put(95.5,7.7){\textcolor{black}{ {\tiny n/a}}}
			\put(95.5,5.7){\textcolor{black}{ {\tiny n/a}}}
			
		\end{overpic}
		\caption{An example of predicted depth hint and the corresponding ground truth, best viewed in color. `pred' and `gt' respectively denote the predicted depth hint and its ground truth, both resized to fit the image for better visualization. The bins of depth hint with no object center falling into are masked out. Details about the bins and the mask are described in Subsection~\ref{loss_functions}. }
		\label{depth_hint_vis}
	}
\end{figure}
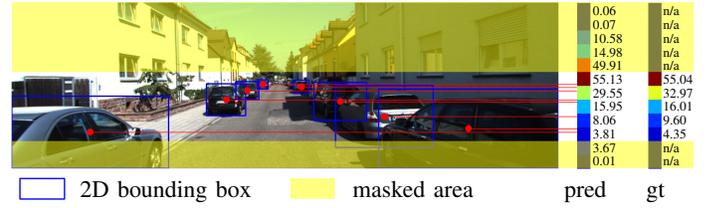

During the training stage, the elements in depth hint vectors are supervised by a depth hint loss, which will be further elaborated on in Subsection~\ref{loss_functions}. Fig.~\ref{depth_hint_vis} shows an example of the predicted depth hint and the corresponding ground truth. It can be seen that the depth hint module is able to learn the depth pattern from the input image, thus providing a reasonable initial guess for subsequent depth estimation.

\subsection{Loss Functions}
\label{loss_functions}
\textbf{Keypoint Loss} Following CenterNet~\cite{zhou2019objects} and CornerNet~\cite{law2018cornernet}, the keypoint branch is trained with a variant of Focal Loss~\cite{lin2017focal}:

\begin{equation}
\label{eq6}
L_{kp}=\frac{-1}{N}\sum_{cuv}\begin{cases}
(1-\hat{y}_{cuv})^\alpha {\rm log}(\hat{y}_{cuv})&{\rm if} \; y_{cuv}=1\\
(1-y_{cuv})^\beta \hat{y}_{cuv}^\alpha {\rm log}(1-\hat{y}_{cuv}) &{\rm otherwise}
\end{cases}
\end{equation}
where $N$ denotes the total number of objects. $y_{cuv}$ and $\hat{y}_{cuv}$ stand for the ground truth and the estimated score for category $c$ at position ($u,v$) in the heatmap. The ground truth for the heatmap is labeled by drawing gaussian kernels at locations that correspond to the projected 3D centroids. Readers are referred to \cite{zhou2019objects} and \cite{law2018cornernet} for more details. The hyperparameters $\alpha, \beta$ are set to $\alpha=2$, $\beta=4$ following CenterNet\cite{zhou2019objects}.

\textbf{Regression Loss}
The outputs from 2D and 3D regression branches are respectively supervised by $L_{{\rm reg2d}}$ and $ L_{{\rm reg3d}} $, both designed to be in a disentangled form inspired by MonoDIS\cite{simonelli2019disentangling}. Let $\psi_{2d}$ be a subset of 2D outputs $\Psi_{2d}= \{ \{ \Delta u_{2d}, \Delta v_{2d} \}, \{ w, h \} \}$ and $-\psi_{2d}$ be its complementary subset. Let $y_{\psi}$ and $\hat{y}_{\psi}$ respectively denote the ground truth values and the estimations of $\psi$, and denote the $L_1$ loss function by $L_1(\cdot) $. The loss function for 2D regression heads is defined as

\begin{equation}
L_{{\rm reg2d}}(y,\hat{y})=\!\!\!\!\!\!\sum_{\psi_{2d}\in \Psi_{2d}}\!\!\!\!\!\!L_1(\mathcal{F}(y_{\psi_{2d}},y_{-\psi_{2d}}),\mathcal{F}(\hat{y}_{\psi_{2d}},y_{-\psi_{2d}}))
\end{equation}

with

\begin{equation}
\mathcal{F}:= \begin{bmatrix}
u_{2d} \\
v_{2d}
\end{bmatrix} + \begin{bmatrix}
\pm w/2\\
\pm h/2
\end{bmatrix}
\end{equation}
denoting the encoding transformation of 2D bounding boxes.

Likewise, by defining $\Psi_{3d}= \{ \{ {\rm cos}\alpha, {\rm sin}\alpha \}, \{ \Delta H,\Delta W,\\ \Delta L \}, \{\Delta u_{3d}, \Delta v_{3d}\}, \{ \widetilde{d}\} \}$, we have
\begin{equation}
L_{{\rm reg3d}}(y,\hat{y})=\!\!\!\!\!\!\sum_{\psi_{3d}\in \Psi_{3d}}\!\!\!\!\!\!L_1(\mathcal{G}(y_{\psi_{3d}},y_{-\psi_{3d}}),\mathcal{G}(\hat{y}_{\psi_{3d}},y_{-\psi_{3d}}))
\end{equation}
with
\begin{equation}
\mathcal{G}:= \bigl( \begin{bmatrix}
x_{3d} \\
y_{3d} \\
z_{3d} \\
\end{bmatrix} + \begin{bmatrix}
\pm W/2 \\
\pm H/2 \\
\pm L/2 \\
\end{bmatrix}\bigr)R
\end{equation}
denoting the encoding transformation of 3D bounding boxes, where $R$ is the rotation matrix of yaw angle $\theta$.

Since the objects with larger depth values generally have smaller 2D sizes, we further adapt the loss function for 2D regression heads to be depth-aware. To be specific, $L_{{\rm reg2d}}$ is remoulded as $d^\gamma L_{{\rm reg2d}}$, where $d$ is the ground truth of depth and hyperparameter $\gamma$ is empirically set to be 0.4 in the experiments.

\textbf{Depth Hint Loss}
A depth hint loss is proposed to explicitly guide the generation of the depth hint vector. Given an input image with a resolution of $ W \times H $, we evenly divide it into $H/32$ bins along the vertical dimension. The depths of objects with 2D centers falling into each bin are accumulated and averaged. The resulted vector with $H/32$ elements serves as the ground truth for the depth hint vector. Let $\hat{\xi}$ be the network's estimation of depth hint vector and $\xi$ be the corresponding ground truth. The depth hint loss is defined as
\begin{equation}
L_{{\rm dh}}=M_{{\rm dh}}L_1(\xi,\hat{\xi})
\end{equation}
where $M_{{\rm dh}}$ is a 0/1 mask which takes 0 if a bin contains no object centers and takes 1 otherwise.

\textbf{Total Loss}
The total loss function is a weighted sum of all the losses discussed above, that is
\begin{equation}
\label{eq12}
L_{{\rm total}}=\boldsymbol{\lambda}[L_{{\rm kp}}, d^\gamma L_{{\rm reg2d}}, L_{{\rm reg3d}}, L_{{\rm dh}}]^T
\end{equation}
where $\boldsymbol{\lambda}=[1, \lambda_1, \lambda_2, \lambda_3]$ is the weighting vector.

\section{Experiments}
\label{experiments}
In this section, the implementation details are firstly introduced in Subsection~\ref{implementation_details}. Quantitative and qualitative experimental results on KITTI are analyzed in Subsection~\ref{evaluation_on_KITTI3D}. To further validate the contributions of this work, ablation study is conducted in Subsection~\ref{ablation_study}. A discussion on the GRU design is given in Subsection~IV-D.
\subsection{Implementation Details}
\label{implementation_details}
\textbf{Experimental Settings}
The performance of the proposed method is evaluated on the KITTI benchmark, which contains a training set with 7481 images and a test set with 7518 images. The detected objects are categorized into \textit{Easy}, \textit{Moderate}, and \textit{Hard} according to the objects' 2D height, occlusional rate, and truncational rate. Since the ground truth labels for the test set are not made publicly available to discourage deliberate network tuning, we split the training set into a training subset with 3712 images and a validation subset with 3769 images following 3DOP~\cite{chen20153d} and conduct ablation study only on the validation subset.

In both training and testing stages, input images are resized into the resolution of (1280,384). We randomly augment the input data by \romannumeral1) flipping with a probability of 0.5 and \romannumeral2) rescaling and shifting with a probability of 0.2. An Adam~\cite{kingma2014adam} optimizer is adopted for updating network's parameters. Four NVIDIA GTX 1080Ti GPUs are used to jointly train our network with a batch size of 32. In all experiments, the values of the hyperparameters involved in Eq.~\ref{eq6} and Eq.~\ref{eq12} are set as in Table~\ref{hyperparameter}.
\begin{table}[htbp]
	\small
	\centering
	\caption{Values of Hyperparameters}
	\label{hyperparameter}
	
	\begin{tabular}{ccccccc}
		\toprule
		Parameter&
		$\alpha$&$\beta$&$\gamma$&$\lambda_1$&$\lambda_2$&$\lambda_3$   \\
		
		\midrule
		Value&
		2&4&0.4&5&2&1\\
		\bottomrule	
	\end{tabular}%
	
\end{table}

\begin{table*}[ht]
	\small
	\centering
	\caption{Comparison with other methods on the test set (IoU=0.7) of KITTI benchmark, measured by ${\rm AP|_{R_{40}}}$. E, M, H are short for \textit{Easy}, \textit{Moderate}, \textit{Hard} respectively.}
	\label{3D_test}
	\begin{tabular}{cccccccccccccc}
		\toprule
		\multirow{2}{*}{Method} & \multirow{2}{*}{Time~(s)} &\multicolumn{3}{c}{${\rm AP_{3D}}$} & \multicolumn{3}{c}{${ \rm AP_{BEV}}$} & \multicolumn{3}{c}{${\rm AP_{2D}}$} & \multicolumn{3}{c}{${ \rm AOS}$}  \\
		&&E&M&H&E&M&H&E&M&H&E&M&H\\ \midrule
		ROI-10D~\cite{manhardt2019roi}&0.2&4.32&2.02&1.46&9.78&4.91&3.74&76.56&70.16&61.15&75.32&68.14&58.98\\
		Shift R-CNN~\cite{naiden2019shift}&0.25&6.88&3.87&2.83&11.84&6.82&5.27&94.07&88.48&78.34&93.75&87.47&77.19\\
		MonoFENet~\cite{bao2019monofenet}&0.15&8.35&5.14&4.10&17.03&11.03&9.05&91.68&84.63&76.71&91.42&84.09&75.93\\
		MonoGRNet~\cite{qin2019monogrnet}&0.04&9.61&5.74&4.25&18.19&11.17&8.73&88.65&77.94&63.31&-&-&-\\
		MonoDIS~\cite{simonelli2019disentangling}&0.10&11.06&7.60&6.37&18.45&12.58&10.66&94.96&89.22&80.58&-&-&-\\
		
		MonoPair~\cite{chen2020monopair}&0.06&13.04&9.99&8.65&19.28&\textbf{14.83}&\textbf{12.89}&\textbf{96.61}&\textbf{93.55}&\textbf{83.55}&91.65&86.11&76.45\\
		SMOKE~\cite{liu2020smoke}&0.03&14.03&9.76&7.84&20.83&14.49&12.75&93.21&87.51&77.66&92.94&87.02&77.12\\
		RTM3D~\cite{li2020rtm3d}&0.05&14.41&\textbf{10.34}&\textbf{8.77}&19.17&14.20&11.99&91.82&86.93&77.41&91.75&86.73&77.18\\
		M3D-RPN~\cite{brazil2019m3d}&0.16&14.76&9.71&7.42&21.02&13.67&10.23&89.04&85.08&69.26&88.38&82.81&67.08\\
		
		Ours~(FADNet)&0.04&\textbf{16.37}&9.92&8.05&\textbf{23.00}&14.22&12.56&96.15&90.49&80.71&\textbf{95.89}&\textbf{89.84}&\textbf{79.98} \\
		\bottomrule
	\end{tabular}%
\end{table*}

\begin{table*}[ht]
	\small
	\centering
	\caption{Comparison with other methods on the validation set (IoU=0.7/IoU=0.5) of KITTI benchmark, measured by ${\rm AP|_{R_{11}}}$.}
	\label{3D_val}
	\begin{tabular}{cccccccccccccc}
		\toprule
		\multirow{2}{*}{Method} & \multirow{2}{*}{Time~(s)} &\multicolumn{3}{c}{${\rm AP_{3D} (IoU=0.7)}$} & \multicolumn{3}{c}{${ \rm AP_{BEV} (IoU=0.7)}$} & \multicolumn{3}{c}{${\rm AP_{3D} (IoU=0.5)}$} & \multicolumn{3}{c}{${ \rm AP_{BEV} (IoU=0.5)}$}  \\
		&&E&M&H&E&M&H&E&M&H&E&M&H\\ \midrule
		CenterNet~\cite{zhou2019objects}&0.04&0.89&1.13&1.35&3.54&4.32&3.57&19.55&18.62&16.61&34.04&30.33&26.78\\
		Deep3DBox~\cite{mousavian20173d}&-&5.85&4.10&3.84&9.99&7.71&5.30&27.04&20.55&15.88&30.02&23.77&18.83\\
		ROI-10D~\cite{manhardt2019roi}&0.2&9.61&6.63&6.29&14.50&9.91&8.73&-&-&-&-&-&-\\
		Shift R-CNN~\cite{naiden2019shift}&0.25&13.84&11.29&11.08&18.61&14.71&13.57&-&-&-&-&-&-\\
		
		MonoGRNet~\cite{qin2019monogrnet}&0.04&13.88&10.19&7.62&-&-&-&50.51&36.97&30.82&-&-&-\\
		Mono3D++~\cite{he2019mono3d++}&0.6&10.60&7.90&5.70&16.70&11.50&10.10&42.00&29.80&24.20&46.70&34.30&28.10\\
		SMOKE~\cite{liu2020smoke}&0.03&14.76&12.85&11.50&19.99&15.61&15.28&-&-&-&-&-&-\\
		MonoFENet~\cite{bao2019monofenet}&0.15&17.54&11.16&9.74&\textbf{30.21}&20.47&17.58&59.93&\textbf{42.67}&\textbf{37.50}&\textbf{66.43}&47.96&43.73\\
		MonoDIS~\cite{simonelli2019disentangling}&0.10&18.05&14.98&13.42&24.26&18.43&16.95&-&-&-&-&-&-\\
		M3D-RPN~\cite{brazil2019m3d}&0.16&20.27&\textbf{17.06}&15.21&25.94&21.18&17.90&48.96&39.57&33.01&55.37&42.49&35.29\\
		RTM3D~\cite{li2020rtm3d}&0.05&20.77&16.86&\textbf{16.63}&25.26&22.12&\textbf{20.91}&54.36&41.90&35.84&57.47&44.16&42.31\\
		
		\midrule
		
		Baseline&0.03&13.61&10.97&9.84&19.73&13.66&12.30&46.42&33.77&31.08&52.91&36.42&31.31 \\
		
		Ours~(FADNet)&0.04&\textbf{23.98}&16.72&15.79&28.70&\textbf{22.24}&19.80&\textbf{60.81}&41.19&35.27&63.24&\textbf{48.12}&\textbf{44.56} \\
		\bottomrule
	\end{tabular}%
\end{table*}

\begin{figure*}[ht]
	{\centering
		\begin{overpic}[width=0.8\textwidth]{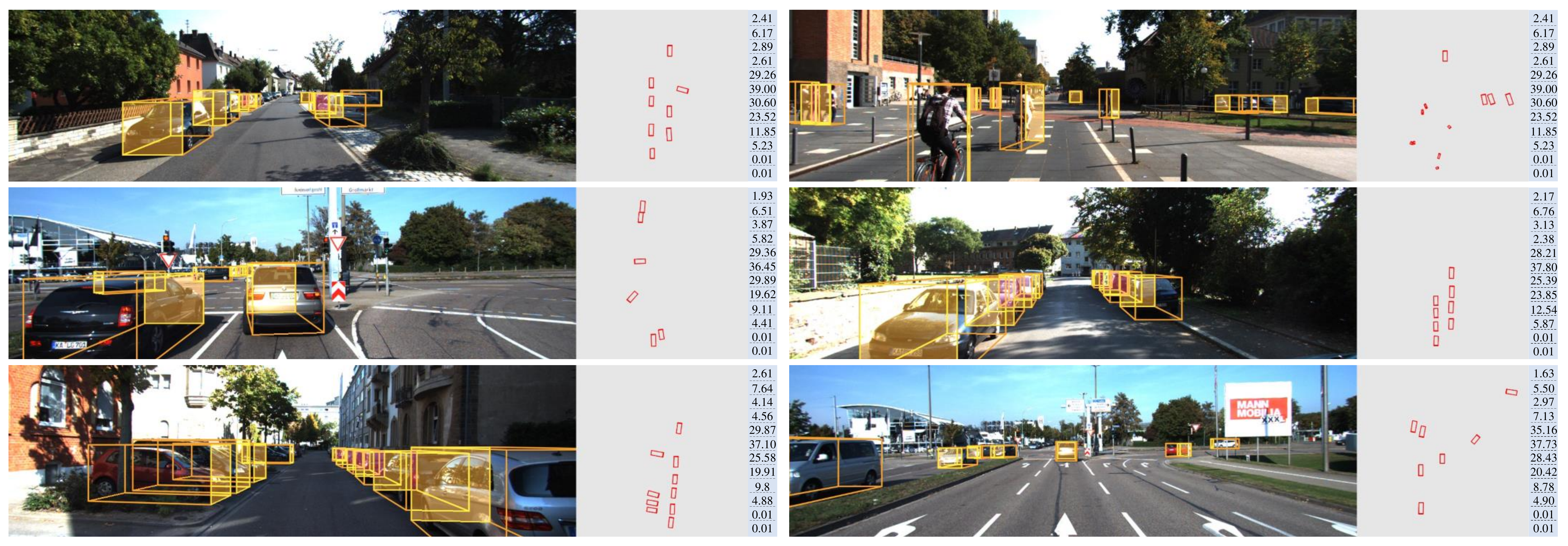}	
		\end{overpic}
		\caption{Qualitative results on KITTI test set, best viewed in color. The predicted depth hint vectors are shown at the rightmost side of each image.}
		\label{qualitative_test}
	}
\end{figure*}

\textbf{Stage-wise Training}
By experiments, it is found that the single-shot training strategy is harmful to the performance due to difficulty in reconciling the gradients from different parts of the network. Therefore, a stage-wise scheme is leveraged for network training. In specific, the network is trained in three stages:

\romannumeral1) Train for 90 epochs with an initial learning rate of 2e-4, which drops by half every 30 epochs.

\romannumeral2) Fix the parameters of depth hint module and accordingly exclude the depth hint loss from the total loss function. Then train for 30 epochs with a learning rate of 3e-5.

\romannumeral3) Further fix the parameters of heatmap head and 2D regression head and accordingly exclude keypoint loss and 2D regression loss from the total loss function. Then train for 30 epochs with a learning rate of 3e-5 and a weight decay of 1e-5.

\textbf{Runtime Testing}
The presented network is implemented in the Pytorch~\cite{PyTorch} programming framework. The batch size of input image is set to 1 during the test stage. Being tested on a personal computer with an Intel i7-9700 CPU and a NVIDIA RTX 2080Ti GPU, the FADNet runs at a speed of 24.3 FPS. Since the running time randomly varies in different runs, the reported speed is calculated by averaging the speeds of 5 runs.

\subsection{Experiments on KITTI}
\label{evaluation_on_KITTI3D}
\textbf{Quantitative Analysis}
To evaluate the performance of the proposed method, the evaluation results on both the test set (Table~\ref{3D_test}) and the validation set (Table~\ref{3D_val}) of KITTI benchmark are reported. Following the convention of most previous works, all the comparisons are made with regard to the `car' class.

For the test set, metrics including the average precision of 3D bounding boxes ${\rm AP_{3D}}$, the average precision of bird's eye view ${\rm AP_{BEV}}$, the average precision of 2D bounding boxes ${\rm AP_{2D}}$, and the average orientation similarity AOS, are employed for a comprehensive evaluation. The commonly used IoU threshold of 0.7 is adopted. Following the proposal of MonoDIS~\cite{simonelli2019disentangling}, all metrics are measured by the 40-point Interpolated Average Precision (${\rm AP|_{R_{40}}}$) scores. As is shown in Table~\ref{3D_test}, the presented FADNet performs competitively against state-of-the-art methods. Notably, our FADNet outperforms the other methods by a large margin in view of AOS, and ranks the second best in view of ${\rm AP_{2D}}$. As for the \textit{Easy} category in 3D metrics, FADNet achieves the highest score on ${\rm AP_{3D}}$ and ${\rm AP_{BEV}}$. Compared with the state-of-the-art method M3D-RPN~\cite{brazil2019m3d} which adopts the region proposal-based framework, the proposed FADNet not only increases the ${\rm AP_{3D}}$ score by 1.61 and the ${\rm AP_{BEV}}$ score by 1.98 for the \textit{Easy} category, but also runs around 4 times faster.

For the validation set, metrics including ${\rm AP_{3D}}$ and ${\rm AP_{BEV}}$ with IoU thresholds of 0.7 and 0.5 are employed for evaluation. The validation metrics are measured by ${\rm AP|_{R_{11}}}$ scores since the ${\rm AP|_{R_{40}}}$ scores are not reported by many of the works. An ablated version of our network, i.e. designed without sequential feature association and depth hint augmentation, is adopted as the baseline for our method. As is shown in Table~\ref{3D_val}, the presented FADNet is superior to the baseline with regard to all metrics and performs on par with other state-of-the-art methods.

\begin{figure*}[ht]
	{\centering
		\begin{overpic}[width=0.8\textwidth]{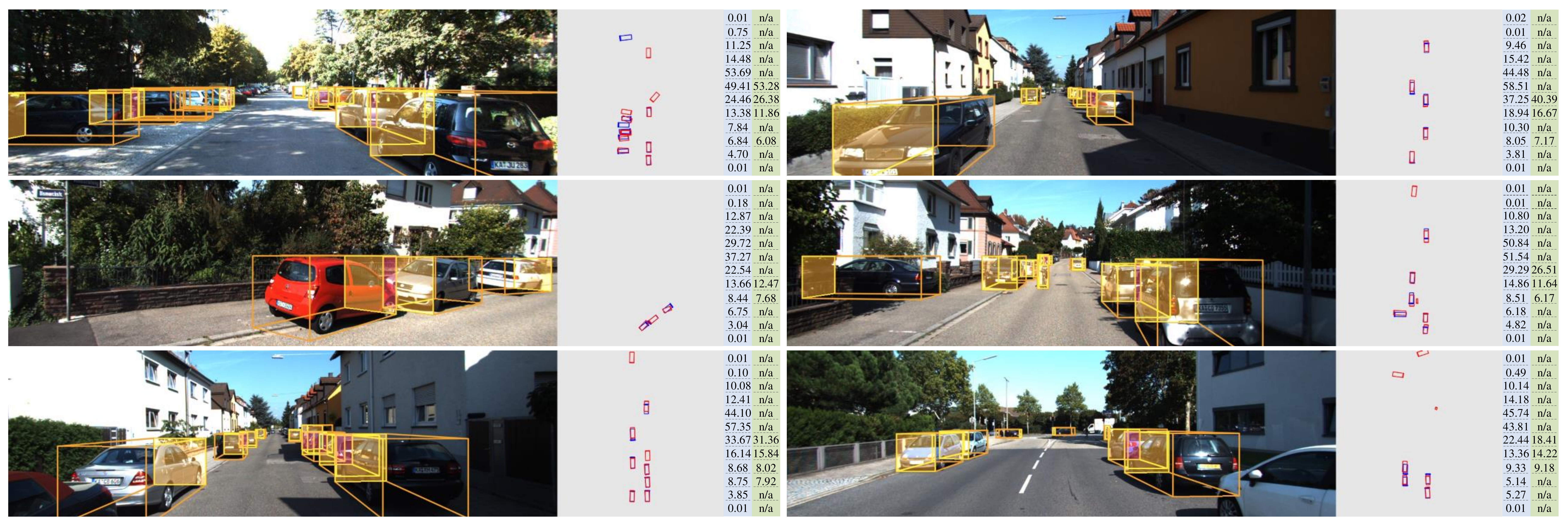}	
		\end{overpic}
		\caption{Qualitative results on KITTI validation set, best viewed in color. For the bird's eye view, the ground truth labels are colored in blue and the network's estimations are colored in red. For the depth hint vectors, the prediction and ground truth are masked in blue and green respectively. (The same goes for Fig. 11.)}
		\label{qualitative_val}
	}
\end{figure*}

\textbf{Qualitative Analysis}
For an intuitive understanding of our method, some visualized detection results on KITTI test set and validation set are respectively exhibited in Fig.~\ref{qualitative_test} and Fig.~\ref{qualitative_val}. The visualization for bird's eye view and the predicted depth hint vectors (unit: ${\rm m}$) are concatenated to the right side of each image. For intuitive comparison, the corresponding ground truth labels are also presented for the examples from validation set. Note that objects not satisfying either of the \textit{Easy}, \textit{Moderate}, \textit{Hard} standards of KITTI (e.g. objects that are too far away, heavily occluded, or heavily truncated) are not plotted. From the exhibited qualitative results, it can be seen that the proposed FADNet is able to detect objects of different scales with decent accuracy.

\begin{figure*}[ht]
	{\centering
		\begin{overpic}[width=0.8\textwidth]{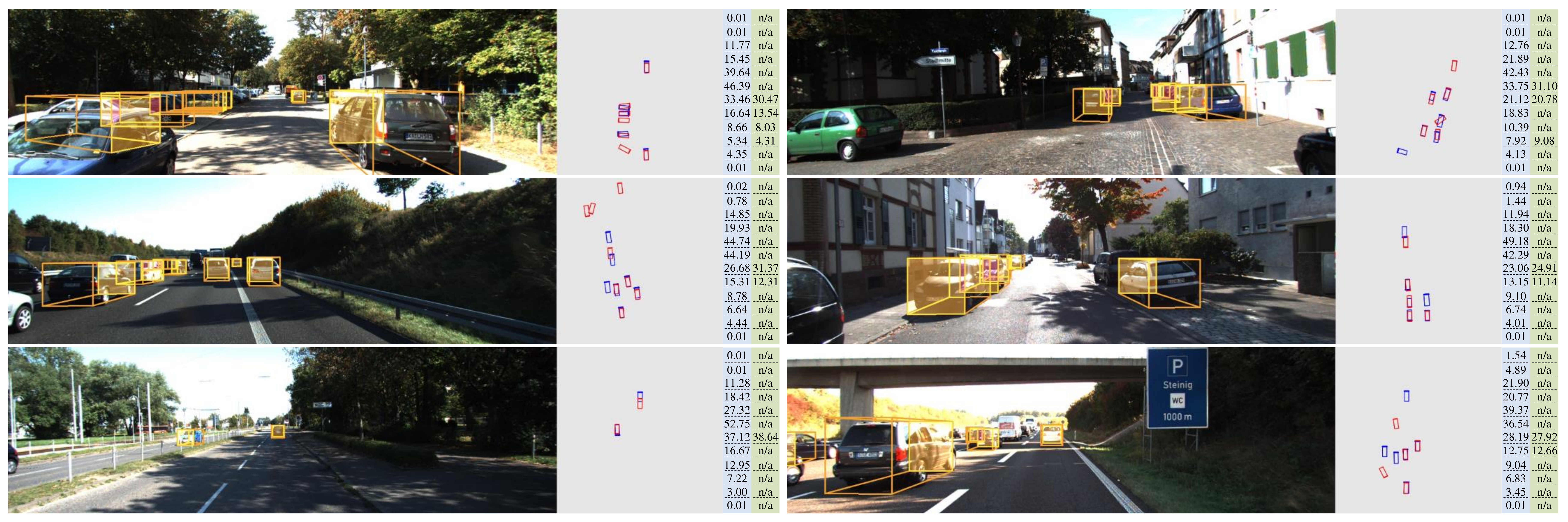}	
		\end{overpic}
		\caption{Examples of failure cases on KITTI validation set, best viewed in color.}
		\label{failures}
	}
\end{figure*}

However, there is still much leeway for improvement. Examples of failure cases produced by the FADNet are shown in Fig.~\ref{failures}. It is observed that detection failures generally happen to three types of cars: \romannumeral1)~Cars in large distance. Monocular 3D detection methods rely on 2D appearance for 3D estimation. For cars in large distance, a large change in depth merely cause a small change in 2D appearance, leading to the difficulty in depth estimation. \romannumeral2)~Cars with large occlusional rates, usually in crowded scenes. \romannumeral3)~Cars with large truncational rates. We will be devoted to researching on these issues in future work.


\subsection{Ablation Study}
\label{ablation_study}
To further evaluate the two initiatives of this work, six variants of FADNet are devised for ablation study, including \romannumeral1)~Baseline, by removing the convGRU and the depth hint module from FADNet. The feature extracted by the backbone network is directly fed into the keypoint head and regression heads. \romannumeral2)~Baseline+FA, by only discarding the depth hint module. \romannumeral3)~Baseline+DH, by only abandoning the sequential feature association. \romannumeral4) Baseline+ConvSep, by replacing the convGRU with four sequentially connected convolutional blocks. \romannumeral5) Baseline+ConvOne, by replacing the convGRU with a single convolutional block. The channels of the convolutional blocks in \romannumeral4) and \romannumeral5) are adjusted such that the total amount of parameters is similar to the original FADNet. \romannumeral6)~Reversed, by reversing the order of sequential feature association. The 2D and 3D evaluation results on KITTI are respectively exhibited in Table~\ref{ablation_2d} and Table~\ref{ablation_3d}. It should be noted that the performance exhibited here is measured by ${\rm AP|_{R_{40}}}$ for a more accurate evaluation, so that it is numerically different from that in Table~\ref{3D_val}.

\begin{table}[ht]
	\small
	\centering
	\caption{Ablation study in 2D metrics on KITTI validation set (IoU=0.7) with different variants of FADNet, measured by ${\rm AP|_{R_{40}}}$. `Base' is short for `Baseline' due to the space limit.}
	\label{ablation_2d}
	
	\begin{tabular}{ccccccc}
		\toprule
		\multirow{2}{*}{Method} &\multicolumn{3}{c}{${\rm AP_{2D}}$} & \multicolumn{3}{c}{${ \rm AOS}$} \\
		&E&M&H&E&M&H\\
		
		\midrule
		Baseline&96.63&89.61&80.65&95.90&88.31&78.15\\
		Base+FA&97.60&90.55&82.88&\textbf{97.68}&89.65&81.70\\
		Base+DH&97.45&89.67&81.11&95.34&88.27&79.52\\
		Base+ConvSep&96.14&88.58&80.22&95.14&89.11&79.92\\
		Base+ConvOne&96.93&89.72&81.29&96.46&90.53&81.38\\
		Reversed&\textbf{98.13}&\textbf{91.24}&\textbf{83.57}&97.62&91.03&81.83\\
		Base+FA+DH&97.67&89.74&81.38&97.53&\textbf{91.04}&\textbf{82.98}\\
		\bottomrule	
	\end{tabular}%
	
\end{table}

As is shown in Table~\ref{ablation_3d}, ${\rm AP_{3D}}$ and {${ \rm AP_{BEV}}$} are increased by introducing either sequential feature association or depth hint augmentation, which validates the effectiveness of the presented two initiatives. By incorporating both of them (i.e. the original FADNet), the network achieves the best performance compared with the other variants. On the other hand, by only reversing the order of sequential feature association and keeping the other configurations unchanged, the network's performance is inferior to FADNet.

\begin{table}[ht]
	\small
	\centering
	\caption{Ablation study in 3D metrics on KITTI validation set (IoU=0.7) with different variants of FADNet, measured by ${\rm AP|_{R_{40}}}$.}
	\label{ablation_3d}
	
	\begin{tabular}{ccccccc}
		\toprule
		\multirow{2}{*}{Method} &\multicolumn{3}{c}{${\rm AP_{3D}}$} & \multicolumn{3}{c}{${ \rm AP_{BEV}}$} \\
		&E&M&H&E&M&H\\
		
		\midrule
		Baseline& 10.33& 8.52&7.67&18.04&12.95&10.81\\
		Base+FA&17.31&11.71&10.66&22.55&15.57&13.85\\
		Base+DH&12.25&8.98&8.14&19.55&14.12&12.53\\
		Base+ConvSep&14.35&9.91&8.61&20.04&14.31&13.24\\
		Base+ConvOne&15.87&10.93&9.80&21.68&15.02&14.21\\
		Reversed&13.53&9.12&9.21&20.30&13.98&13.02\\
		Base+FA+DH&\textbf{18.40}&\textbf{12.52}&\textbf{10.97}&\textbf{25.55}&\textbf{16.54}&\textbf{15.53}\\
		\bottomrule	
	\end{tabular}%
	
\end{table}

Besides, it can be observed from Table~\ref{ablation_2d} that ${\rm AP_{2D}}$ and {${ \rm AOS}$} are hardly enhanced by the two initiatives. This can be explained from two aspects: \romannumeral1) The features for 2D estimation are generated by the first and second timesteps of convGRU, hence having limited guidance from the other groups. \romannumeral2) 2D information is relatively easy to estimate. Even by reversing the order of feature association (refer to the `Reversed' variant), the hidden states from 3D estimations are limited in refining 2D estimations. Simply put, the hints from harder estimations generally fail to guide the easier estimations.

It should be noted that we build our network on the basis of a very simple baseline in order to emphasize the contributions of the two proposed initiatives. Now that \romannumeral1)~both initiatives have been testified to be effective by the ablation study and \romannumeral2)~both initiatives can be decoupled from the main branch, our method still has the potential to achieve higher metric scores on the KITTI benchmark, if the two initiatives are applied to more complex frameworks or combined together with other techniques such as using depth priors and post optimization.

\subsection{Discussion on the GRU design}
\label{discussion}
A direct motivation of using a GRU instead of a CNN design to associate features is that GRU is more efficient in capturing the dependency between different components. The delicately designed gates inside a GRU module help to decide which part of former information to be forgotten and which part to be retained. For instance, suppose the 2D regression head identifies an object as `a large car'. The keyword `large' suggests that this car is not very distant from us, thereby the regression head for predicting depth is guided by such message flowing from a prior stage.

In fact, the shortcut design of ResNet~\cite{he2016deep}, which enables a direct message flow, implements a similar structure with RNN. However, a GRU is endowed with the extra ability of selectively forgetting information from prior stages. Furthermore, as also discussed in~\cite{liao2016bridging}, although a ResNet can be converted into an RNN form, an RNN can perform on par with a ResNet while considerably reducing the parameter amount. In other words, RNN is more efficient in understanding the sequential association between features, which is coherent with the experimental results exhibited in Subsection~IV-C.
\section{Conclusion}
\label{conclusion}
In this work, a singe-stage keypoint-based monocular 3D object detection network named as FADNet has been presented. On the one hand, convolutional features that correspond to different output modalities have been tackled differently by adopting the strategy of output grouping. A convolutional GRU has been integrated to establish sequential association across output features from different groups. It has been found by experiment that this design contributes to the consistency between 2D and 3D estimations and improves the overall performance on the KITTI benchmark. On the other hand, motivated by the observation that the objects' depths are unevenly distributed along the vertical dimension of input images, a dedicated depth hint module has been devised to estimate the depth pattern. The depth hints output by the depth hint module are capable of providing informative hints for subsequent depth estimation. The presented FADNet performs competitively against state-of-the art methods, providing a good trade-off between detection accuracy and running speed.

\bibliographystyle{IEEEtran}
\bibliography{reference}

\begin{IEEEbiography}[{\includegraphics[width=1in,height=1.25in,clip,keepaspectratio]{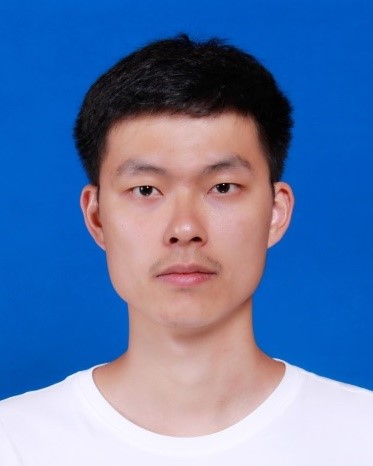}}]{Tianze Gao}
received the B.S. degree in automation from the Harbin Institute of Technology, Harbin, China.
Currently, he is working towards the Ph.D. degree in control science and technology at the Harbin Institute of Technology, Harbin, China. His research interests include autonomous driving technology, computer vision and artificial intelligence theories, with a particular focus on object detection and tracking in autonomous driving scenarios.
\vspace{-25pt}
\end{IEEEbiography}

\begin{IEEEbiography}[{\includegraphics[width=1in,height=1.25in,clip,keepaspectratio]{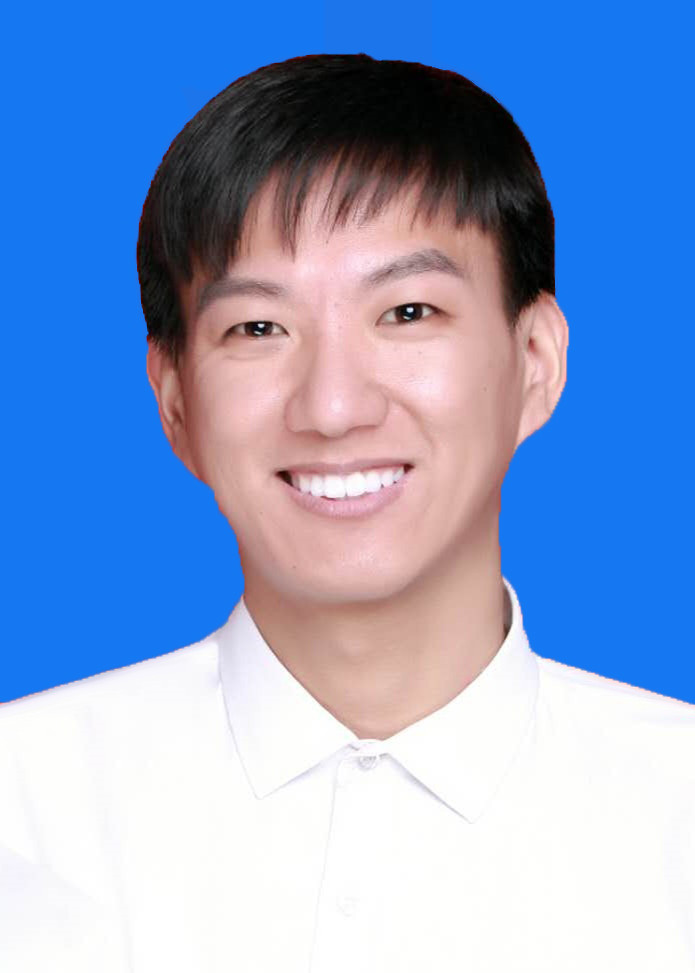}}]{Huihui Pan}
	received the Ph.D. degree in control science and engineering
	from Harbin Institute of Technology, Harbin, China, in 2017, and the Ph.D.
	degree in mechanical engineering from the Hong Kong Polytechnic
	University, Hong Kong, in 2018.

Since December 2017, he has been with
	the Research Institute of Intelligent Control and Systems, Harbin Institute
	of Technology, Harbin, China. His research
interests include nonlinear control, vehicle dynamic
control, and intelligent vehicles.
	\vspace{-25pt}
\end{IEEEbiography}

\begin{IEEEbiography}[{\includegraphics[width=1in,height=1.25in,clip,keepaspectratio]{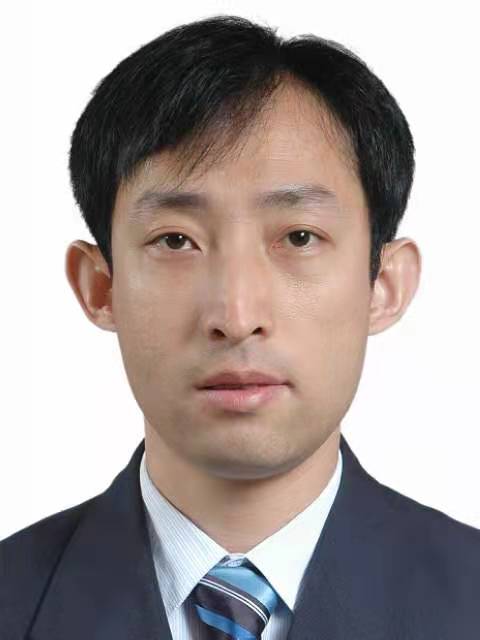}}]{Huijun Gao}
(Fellow, IEEE) received the Ph.D. degree in control science and engineering from Harbin Institute of Technology, Harbin, China, in 2005. From 2005 to 2007, he carried out his post-doctoral research with the Department of Electrical and Computer Engineering, University of Alberta, Edmonton, AB, Canada. Since 2004, he has been with the Harbin Institute of Technology, where he is currently a Chair Professor and the Director of the Research Institute of Intelligent Control and Systems. His research interests include intelligent and robust control, robotics, mechatronics, and their engineering applications.

Dr. Gao is a Vice President of IEEE Industrial Electronics Society and a Council Member of IFAC. He serves/served as Editor-in-Chief of the IEEE/ASME Transactions on Mechatronics, Co-Editor-in-Chief of the IEEE Transactions on Industrial Electronics, and Associate Editor of the Automatica, the IEEE Transactions on Cybernetics, and the IEEE Transactions on Industrial Informatics, etc. He is a Member of Academia Europaea and a Distinguished Lecturer of the IEEE Systems, Man and Cybernetics Society.
\end{IEEEbiography}

\end{document}